\documentclass[journal]{IEEEtran}
\usepackage{amsmath,amsfonts,amssymb}
\usepackage{mathtools}
\usepackage{cite}
\usepackage{algorithm2e}
\usepackage{enumitem}
 \usepackage{xcolor}
\usepackage{pdflscape}
\usepackage{adjustbox}
\usepackage{booktabs}
\usepackage{array}
\begin{document}
\title{An AI-Enabled Hybrid Cyber-Physical Framework for Adaptive Control in Smart Grids}
\author{Muhammad~Siddique~\IEEEmembership{(Member~IEEE),}%
\thanks{Muhammad Siddique is with the Department of Electrical Engineering, NFC Institute of Engineering and Technology (IET), Multan, Pakistan (e-mail: engr.siddique01@gmail.com).}
\and
Sohaib~Zafar%
\thanks{Sohaib Zafar is with the Lahore University of Management Sciences (LUMS), Lahore, Pakistan (e-mail: 17060003@lums.edu.pk).}
}

\markboth{IEEE Transactions on Smart Grid,~Vol.~XX, No.~XX, July~2025}%
{Siddique \MakeLowercase{\textit{et al.}}: Hybrid Cyber-Physical Modeling of Smart Grid for AI-Based Adaptive Control}

\maketitle

\begin{abstract}
Evolving smart grids require flexible and adaptive control methods. A harmonized hybrid cyber-physical framework, which considers both physical and cyber layers and ensures adaptability, is one of the critical challenges to enable sustainable and scalable smart grids. This paper proposes a three-layer (physical, cyber, control) architecture, with an energy management system as the core of the system. Adaptive Dynamic Programming(ADP) and Artificial Intelligence-based optimization techniques are used for sustainability and scalability. The deployment is considered under two contingencies: Cloud Independent and cloud-assisted. They allow us to test the proposed model under a low-latency localized decision scenario and also under a centralized control scenario.   The architecture is simulated on a standard IEEE 33-Bus system, yielding positive results. The proposed framework can ensure grid stability, optimize dispatch, and respond to ever-changing grid dynamics. 
\end{abstract}

\begin{IEEEkeywords}
Smart grid, cyber‑physical model, adaptive control, optimization, demand response, game theory, resilience, reinforcement learning.
\end{IEEEkeywords}

\section*{List of Symbols}

\begin{description}[labelwidth=1cm, labelsep=0.5cm, leftmargin=2cm, style=multiline]
  \item[$t$] Continuous time variable
  \item[$T$] Final time horizon for optimization
  \item[$J$] Objective cost functional
  \item[$u(t)$] Control input vector at time $t$ (e.g., curtailment, charging actions)
  \item[$x(t)$] System state vector at time $t$ (e.g., voltages, SOC)
  \item[$y(t)$] Output vector (measured or controlled outputs)
  \item[$A$, $B$, $C$, $D$] System matrices for linear or linearized state-space model
  \item[$P_G$] Power generated by controllable sources
  \item[$P_L$] Power curtailed from load demand
  \item[$R(t)$] Renewable generation at time $t$
  \item[$L(t)$] Total load demand at time $t$
  \item[$E$] Energy level in energy storage systems (e.g., batteries)
  \item[$S(t)$] State of charge (SOC) of energy storage
  \item[$C_G(P_G)$] Cost of conventional power generation
  \item[$C_L(P_L)$] Cost of load curtailment
  \item[$C_E(E)$] Cost associated with energy storage
  \item[$C_S(u)$] Cost of control action (e.g., switching, actuation)
  \item[$\eta$] Efficiency of energy conversion/storage
  \item[$\lambda$] Lagrange multiplier (for constraints or dual formulations)
  \item[$\delta$] Perturbation or variation (e.g., in adaptive control)
  \item[$\tau$] Time constant or control delay
  \item[$P_{ij}(t)$] Power flow from bus $i$ to bus $j$ at time $t$
  \item[$Z_{ij}$] Impedance of transmission line between nodes $i$ and $j$
  \item[$V_i(t)$] Voltage magnitude at bus $i$
  \item[$\theta_i(t)$] Voltage angle at bus $i$
  \item[$f_i(t)$] Frequency at bus $i$
  \item[$\Delta f(t)$] Frequency deviation from nominal
  \item[$Q(x,u)$] State-action value function (in reinforcement learning)
  \item[$r_t$] Instantaneous reward at time $t$
  \item[$\gamma$] Discount factor for future rewards (RL)
  \item[$\pi$] Policy mapping state to actions in reinforcement learning
  \item[$\hat{x}(t)$] Estimated state (e.g., from observer or filter)
  \item[$N$] Prediction/control horizon in Model Predictive Control (MPC)
  \item[$H$] Hessian matrix in quadratic programming formulation
  \item[$F$] Forecast function for future renewables/load
  \item[$\sigma$] Standard deviation (used in uncertainty modeling)
  \item[$\mathcal{P}$] Set of prosumers or controllable participants
  \item[$\mathcal{N}$] Set of all network nodes (buses)
  \item[$\mathcal{E}$] Set of all network edges (lines)
  \item[$B_{ij}$] Susceptance between bus $i$ and $j$
  \item[$u_i^{\text{max}}, u_i^{\text{min}}$] Upper and lower control bounds
  \item[$E_{\text{max}}, E_{\text{min}}$] Max/min energy limits for storage systems
  \item[$P_G^{\text{max}}$] Maximum generation capacity of dispatchable generators
  \item[$D(t)$] Demand profile (possibly stochastic)
  \item[$W(t)$] Weather input (for forecasting solar/wind)
  \item[$k$] Discrete-time index (if applicable)
\end{description}

\section{Introduction}
\label{sec:introduction}

\IEEEPARstart Rising energy demand, urgent need for decarbonization, increasing penetration of stochastic renewable energy sources (RES), and electrification of transport have necessitated the transformation of the traditional electric grid to an intelligent, adaptive, and sustainable smart grid.\cite{fang2012smart, amin2005toward, farhangi2010path}. Advancements in sensors have contributed to fusing the grid with a communication network and computational intelligence. The application of these advancements makes the monitoring, control, and optimization of the process possible in real-time across the layers. The aggregation of such a structure is what we know as a cyber-physical system(CPS). 
 \cite{gungor2011smart, mohsenian2010autonomous, momoh2012smart}.

The system complexity and uncertainty at the operational level have significantly increased due to highly flexible loads (EVs) and integration of distributed renewable energy resources (DERs). \cite{zhang2011game, chakraborty2016power, yang2019distributed}, making traditional model-based control approaches ineffective\cite{liu2009cyber, teixeira2015secure}. Particularly, such centralized architectures often face impediments due to latency, single-point failures, and scalability, reducing their effectiveness for the decentralized structure of the modern smart grids  \cite{gao2012cyber, huang2018review}. 
Therefore, adaptable control architectures with standardized interfaces are warranted. They help resolve the core issue of effective coordination between spatio-temporally dispersed components, since these components may operate under different control regimes. The components may include DER units, microgrids, and EV Charging points.\cite{ieeep2030, ieee1547, yan2013survey}. 

In addition to spatio-temporal decentrality, the monitoring and control solution for the modern grid needs to be responsive to stochastic events. The monitoring and control solution must be responsive to highly stochastic events such as generation intermittency, load fluctuations, and device-level failures \cite{khanna2016energy, palensky2011demand}. Moreover, with the increasing fusion of information and communication technologies, the smart grid faces vulnerability vis-à-vis cyber attacks. Cyber attacks such as False Data Injection(FDI) and denial of service(DoS) have a demonstrated capability to undermine control systems and falsify state estimation \cite{liu2011false, sridhar2012cyber, rawat2015cybersecurity}. Often, these attacks are stealthy and hence compromise the false data detection mechanisms in place, thereby risking the system stability. Lag introduced by the centralized cloud computing platforms is another central issue in the realization of a truly intelligent system. The network delays and resource allocation prove to be limiting the efficacy of otherwise scalable cloud systems. These limitations are especially a problem for time-sensitive grid operations such as voltage control, frequency regulation, and switch protection \cite{alam2017fog, tao2017edge}. Research has been done on edge and fog computing solutions for such problems. These solutions enable localized processing at the edge of the network, thereby improving the response time and resilience of the grid. \cite{kulkarni2019edge, mohagheghi2010distributed}.

Apart from latency, interoperability is also of core concern. While industry standards have been developed to outline guidelines for interfacing and integrating heterogeneous systems \cite{ieeep2030, ieee1547, smartgridinterop}, ensuring compliance throughout the legacy infrastructure and advanced technologies remains a continuing challenge. 

Advancements in artificial intelligence (AI), especially in reinforcement learning and adaptive dynamic programming(ADP), offer a powerful data-driven approach towards finding the solution to complex decision-making problems in the context of increasing uncertainty. These “model-free” approaches have shown promise in direct learning from the environment \cite{liu2019reinforcement, chen2021deep}.

Furthermore,agent-based modeling (ABM) and game-theoretic paradigms present robust solutions to represent the behaviors of the grid participants, helping decentralized control and optimization within local and global constraints \cite{saad2012game, sharma2018game}. However, incorporating AI-based methods within a harmonized hybrid CPS framework, which integrates both cyber and physical layers while ensuring adaptability and security, remains a research gap. Furthermore, limited work has been done on hybrid control schemes focused on combining localized edge intelligence with cloud-based optimization. 

This paper proposes an \textit{AI-enabled hybrid cyber–physical framework for adaptive control in smart grids}, seamlessly integrating agent-based modeling, reinforcement learning, and game theory using a layered architecture.  The framework is based on two adaptive control methodologies: (1) a hybrid edge–cloud ADP controller, where local agents perform near-optimal control using local state estimations, while cloud-based agents refine global value functions using strategic coordination; and (2) a multi-agent deep reinforcement learning (DRL) scheme based on Proximal Policy Optimization (PPO) and Deep Q-Networks (DQN), enabling agents to learn from interaction and adapt policies online in a model-free manner.

The architecture is augmented with a model for \textit{cyber-physical resilience}. It includes FDI attack modeling and a resilience index, which quantifies the system’s stability under extreme conditions. The modeling is contextualized in a hybrid CPS, mimicking the interaction and interplay of communication infrastructure, electrical devices, and user behavior and environmental uncertainty.  This framework is validated through extensive simulations on the IEEE 33-Bus radial distribution system, connected with DERs, smart loads under multiple scenarios. The model is evaluated under normal operational conditions, load variability, and generation intermittency, as well as under cyber attacks through FDI. Compared to baseline models, the hybrid approach demonstrated better performance across metrics such as control cost, response time, system stability, and resilience index. 

\textbf{Contributions:} The main contributions of this paper are summarized as follows:
\begin{itemize}
    \item We propose a unified, hybrid CPS framework for smart grid control that integrates agent-based modeling, ADP, and DRL to enable adaptive and decentralized decision-making.
    \item A novel adaptive control algorithm is designed for  a hybrid edge–cloud ADP controller, and a DRL-based scheme employing PPO and DQN for real-time control under uncertainty.
    \item A game-theoretic agent interaction model is formulated to enable strategic coordination among distributed agents while preserving individual autonomy.
    \item A cyber-resilience evaluation module is developed, incorporating FDI attack modeling and resilience index computation to assess system vulnerability and recovery capability.
    \item The effectiveness of the proposed framework is demonstrated through detailed simulation studies on a modified IEEE 33-bus test system under multiple operational and adversarial scenarios.
\end{itemize}

The rest of the paper is organized as follows:  Section~\ref{sec:Overview} presents the overview of the proposed hybrid CPS modeling framework. Section~\ref{sec:Game} describes the agent-based and game-theoretic modeling of the smart grid network. Section~\ref{sec:control} details the design of the two adaptive control algorithms. Section~\ref{sec:FDI} formulates the cyber–physical resilience metrics and FDI model. Section~\ref{sec:Simulation} provides the simulation results and performance analysis. Finally, Section~\ref{sec:Conclusion} concludes the paper with key insights and future directions.

\section{System Overview}
\label{sec:Overview}
This section describes the components constituting our model of the smart grid. A formal tuple is used to model the cyber-physical nature of the smart grid, as given below:

\[
\mathcal{G}_{SG} = \langle \mathcal{P},\,\mathcal{C},\,\mathcal{E},\,\mathcal{U},\,\mathcal{O} \rangle
\]

This abstract formulation mimics the character of the modern-day smart grid and enables a holistic analysis of its principal components. The individual elements of the tuple are elaborated as follows: 

\begin{itemize}
    \item[\(\mathcal{P}\):] \textbf{Physical power system} This consists of the components constituting the physical infrastructure, including generation, transmission, and distribution infrastructure behavior, load, and storage. These are modeled with differential algebraic equations governing their fundamental principles. 
    \item[\(\mathcal{C}\):] \textbf{Cyber-communication infrastructure} This component of the smart grid tuple models the communication protocols, smart meters, sensors, PMUs, and the SCADA system.  
    \item[\(\mathcal{E}\):] \textbf{Energy Management System (EMS)} The EMS is employed for generation and dispatch optimization as well as restoration operation of the grid. 
    \item[\(\mathcal{U}\):] \textbf{Control input space} Control input space includes generator setpoints, load shedding, DER control, load shedding and demand response. 
    \item[\(\mathcal{O}\):] \textbf{Objective space} The objective space is the solution space which includes the performance goals, including cost, reliability, emissions, and grid-resilience. 
\end{itemize}

The tuple-based abstract modeling approach enables us to integrate AI-based adaptive control and reinforcement learning methodologies to ensure grid stability and resilience, efficient resource usage, and scalability of the model. 
\subsection{Physical Layer}

We comprehensively model the physical layer, consisting of the electrical infrastructure, including generation units, transmission, and distribution, as well as the network behavior. The physical layer is modeled for both steady-state and dynamic behaviors related to grid operation and control. 

\subsubsection{AC Power Flow}

We model the power flow as the basic component of the physical layer using the AC power flow equations below:

\begin{equation}
\begin{aligned}
P_i &= \sum_{j\in\mathcal{N}} |V_i||V_j|\bigl(G_{ij}\cos\theta_{ij} + B_{ij}\sin\theta_{ij}\bigr),\\
Q_i &= \sum_{j\in\mathcal{N}} |V_i||V_j|\bigl(G_{ij}\sin\theta_{ij} - B_{ij}\cos\theta_{ij}\bigr),
\end{aligned}
\end{equation}

Active power is quantitatively represented by  (\(P_i\)) and reactive power injection at each bus is given by(\(Q_i\)).  \(|V_i|\)represent the voltage magnitudes and, \(\theta_i\)represents angles respectively. Network admittance is given by  \(G_{ij}, B_{ij}\). 
The solution of the power flow equations provides the following: 

\begin{itemize}
    \item Voltage regulation and limit checks
    \item Optimal power dispatch
    \item Grid contingency analysis
\end{itemize}

AC power flow  being included in the model is the foundation for larger optimization in the smart grid. Optimal Power Flow(OPF) and SCOPF stem from these basic equations. 

\begin{figure}[!h]
    \centering
    \includegraphics[width=0.48\textwidth]{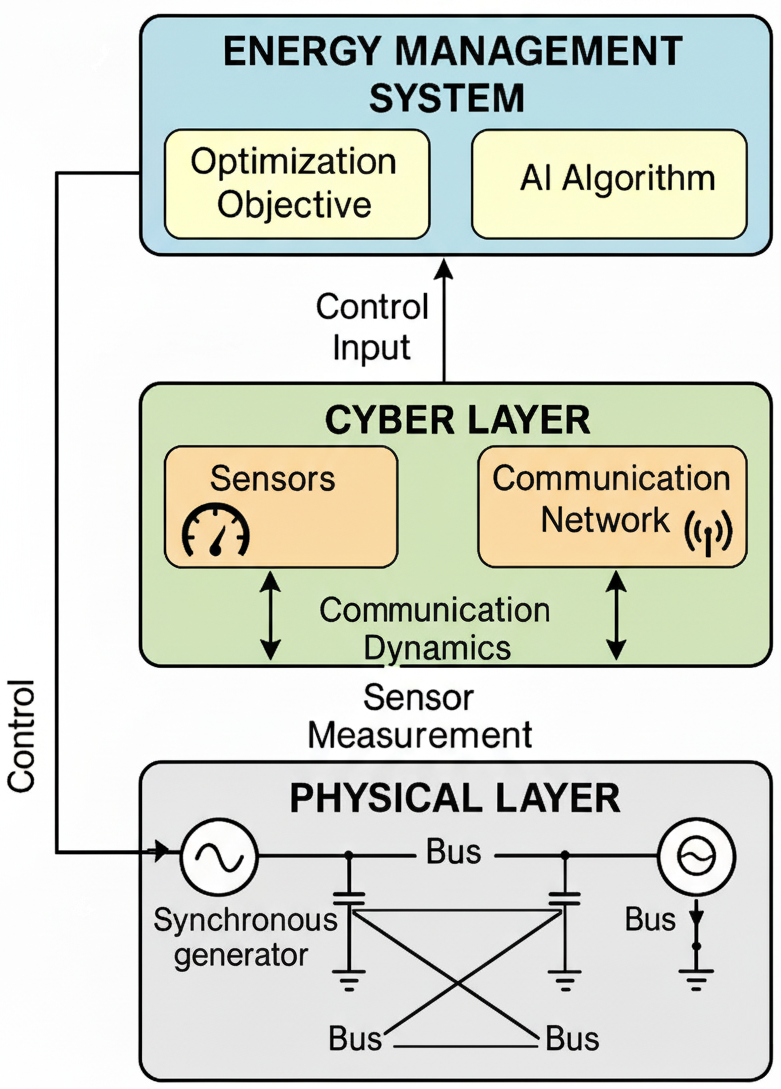}
    \caption{Architecture of hybrid cyber-physical modeling of smart grid for AI-based adaptive control.}
    \label{fig:hybrid_ai_cps}
\end{figure}

\subsubsection{Generator Dynamics}

We model synchronous generator dynamics using the classical swing equation:
\begin{equation}
M_i\ddot\delta_i + D_i\dot\delta_i = P_{m,i} - P_{e,i}
\label{eq:swing}
\end{equation}
where \(\delta_i\) is the rotor angle of generator \(i\), \(M_i\) is the inertia constant, \(D_i\) is the damping coefficient, and \(P_{m,i}\) and \(P_{e,i}\) denote the mechanical input and electrical output power, respectively.

% This foundational equation describes the dynamic electromechanical behavior of the rotor. It is central to the analysis of the system’s response to disturbances. 

% The model allows for analysis of frequency by characterizing rotor speed deviations under power imbalances in the system. Further, it enables stability evaluation employing transient and small-signal analysis. Moreover, the swing equation helps in the design and evaluation of advanced control schemes such as power system stabilizers (PSS), Wide-Area Damping Controllers (WADC), and model predictive control (MPC). In addition, the swing equation helps in the impact analysis of high penetration of inverter-based generation resources (IBR), and the resultant system inertia. 

% In order for a realistic characterization of power system behavior, it is coupled with dynamic models of the excitation system, turbine-governor, and network constraints modeled with non-linear algebraic formulations as described in the next sections. 

\paragraph{Excitation System:}  
The excitation system involves the regulation of the generator’s internal voltage \(E'_{q,i}\) through the field voltage \(E_{fd,i}\). The field voltage affects the electromotive force (EMF) of the generator, which is directly factored in the electrical power output, \(P_{e,i}\) of the generator via the network. This interconnection of the variables couples the electrical and mechanical behavior of the machine.  It is represented through a simplified model given by: 
\begin{equation}
T_{A,i} \dot{E}_{fd,i} = -E_{fd,i} + K_{A,i}(V_{ref,i} - V_i)
\label{eq:exciter}
\end{equation}
where \(T_{A,i}\) is the time constant of the automatic voltage regulator (AVR), \(K_{A,i}\) is the AVR gain, \(V_{ref,i}\) is the voltage reference, and \(V_i\) is the terminal voltage magnitude of the generator.

\paragraph{Turbine-Governor System}
A turbine governor is used to control the input mechanical power according to primary frequency control, which essentially bridges the mechanical dynamics to the frequency dynamics of the power system. The model for the turbine-governor system represents this relationship between mechanical power \(P_{m,i}\) in response to frequency deviations. A first-order differential equation to model this is given as:   
\begin{equation}
T_{g,i} \dot{P}_{m,i} = -P_{m,i} + P_{ref,i} - R_i^{-1} (\omega_i - \omega_0)
\label{eq:governor}
\end{equation}
In this model, \(T_{g,i}\) is the governor time constant, \(P_{ref,i}\) is the mechanical power setpoint, \(R_i\) is the droop coefficient, \(\omega_i = \dot{\delta}_i\) is the rotor speed, and \(\omega_0\) is the nominal system speed. 

\paragraph{Algebraic Network Constraints}  

The nonlinear algebraic power flow equations are representative of the power balance at each bus and characterize all the generators and loads in a network. The electrical power output \(P_{e,i}\) is computed from network conditions through these nonlinear algebraic power flow equations. These are given by:
\begin{align}
P_{e,i} &= \sum_{j \in \mathcal{N}} V_i V_j \left( G_{ij} \cos(\theta_i - \theta_j) + B_{ij} \sin(\theta_i - \theta_j) \right) \label{eq:pe} \\
Q_{e,i} &= \sum_{j \in \mathcal{N}} V_i V_j \left( G_{ij} \sin(\theta_i - \theta_j) - B_{ij} \cos(\theta_i - \theta_j) \right) \label{eq:qe}
\end{align}
Here, \(V_i\) and \(\theta_i\) are the voltage magnitude and phase angle at bus \(i\), \(G_{ij}\) and \(B_{ij}\) are the conductance and susceptance components of the bus admittance matrix \(Y_{bus}\), and \(\mathcal{N}\) is the set of all buses. These equations enforce power balance at each bus and couple the electrical behavior of all generators and loads in the network.

\subsubsection{Full DAE System:}  
The complete system is described by a set of differential-algebraic equations (DAEs) of the form:
\begin{align}
\dot{x} &= f(x, y) \label{eq:dae-dyn} \\
z &= g(x, y) \label{eq:dae-alg}
\end{align}
In this formulation, \(x\) includes the dynamic state variables such as \(\delta_i\), \(\omega_i\), \(E_{fd,i}\), and \(P_{m,i}\), while \(y\) includes algebraic variables such as bus voltage magnitudes \(V_i\) and angles \(\theta_i\). The function f(x,y) is a set of differential equations consisting of swing equation \eqref{eq:swing}, excitation system \eqref{eq:exciter} and turbine-governor model \eqref{eq:governor}. The function g(x, y) represents the algebraic constraints such as the power flow equations \eqref{eq:pe} and \eqref{eq:qe}. This coupled DAE formulation is basic for simulating and analyzing the inter machine oscillations, dynamic stability and frequency response. It also provides the basis for developing grid control strategies and integrating the renewable energy systems into traditional synchronous-machine-dominated power systems.

\subsection{Cyber-Communication Infrastructure}
This part is modeling the cyber communication infrastructure which is represented by \(\mathcal{C}\) in the tuple \(\mathcal{G}_{SG} = \langle \mathcal{P},\,\mathcal{C},\,\mathcal{E},\,\mathcal{U},\,\mathcal{O} \rangle\) . The cyber-communication infrastructure is comprised of the sensors, computation, communication, and information processing modules. In the interaction with the physical layer of the power system, the cyber-communication infrastructure ensures secure and trustworthy monitoring and control. 

To elaborate, the cyber-communication layer includes technologies and components such as phasor measurement units (PMUs), smart meters, SCADA systems, intelligent electronic devices(IEDs), and edge computing infrastructure. 

\subsubsection{Sensor Measurement Models}

% Accurate and timely measurement of system states is essential for ensuring observability, stability, and control in smart grids. The output from a typical sensor \(i\) at time \(t\) is mathematically modeled as:

% \subsubsection{Communication and Information Dynamics}

The cyber-physical power system is highly influenced by the communication network, which interconnects cyber nodes such as substations, DER controllers, and EMS. This network allows for centralized operation as well as local autonomy. The dynamic states of the cyber nodes are affected by constraints introduced by the network limitations, such as packet losses, bandwidth constraints, etc. 

\begin{figure}[!h]
    \centering
    \includegraphics[width=0.4\textwidth]{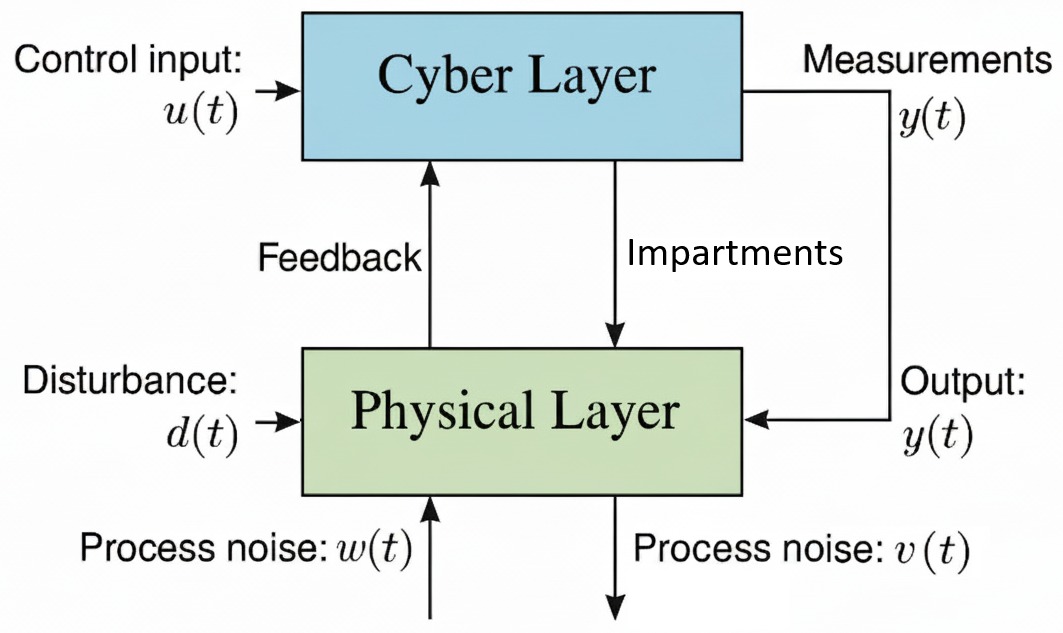}
    \caption{Illustration of the cyber-physical integration in a smart grid, showing interaction between the physical power system and the cyber infrastructure.}
    \label{fig:cps_integration}
\end{figure}

To model the evolution of internal cyber-states under these conditions, we adopt a communication-aware nonlinear state-space formulation:

\begin{equation}
\begin{aligned}
\dot{\hat{x}}_i(t) &= f_i\Big(\hat{x}_i(t), u_i(t - \tau_{ij}(t)), d_i(t)\Big) + B_{w_i} w_i(t), \\
y_i(t) &= h_i\big(\hat{x}_i(t)\big) + v_i(t),
\end{aligned}
\label{eq:cyber_comm_dynamics}
\end{equation}

Here, the  \(\hat{x}_i(t) \in \mathbb{R}^{n_i}\) represents the internal cyber-state vector of node  \(i\). Node includes the relevant digital variables relevant to the local control logic. Further,the term \(u_i(t - \tau_{ij}(t)) \in \mathbb{R}^{m_i}\) mimics the control input from node \(j\).This control input is affected by time-verying communication lag, represented in our formulation as  \(\tau_{ij}(t)\), capturing latency induced due to congestion or asynchronous scheduling.Moreover,  \(d_i(t) \in \mathbb{R}^{p_i}\) mimics the exogenous disturbances such as protocol jitter, system noise, and time drift at  \(i\). Next,  \(w_i(t)\) models communication-induced uncertainties, such as jitter, variable latency, or cyber attacks. \(v_i(t)\)represents measurement noise and digital corruption in sensor and control signals. The matrix \(B_{w_i}\) governs how such disturbances propagate into the cyber-state dynamics of node \(i\). Lastly, output vector \(y_i(t) \in \mathbb{R}^{q_i}\)  , includes raw or processed data streams, timestamps, delay aknowledgements or other observable communication metrics at node \(i\).

Furthermore, \(f_i(\cdot)\) models the evolution of control law implementation, local optimization routines, and protocol stack processing \(i\). The function \(h_i(\cdot)\) mimics the output logic of the node. This includes sensing feedback, diagnostic responses, and actuation signals.

The communication-induced phenomena of the modern smart-grid environment are given by a comprehensive framework represented in Equation~\eqref{eq:cyber_comm_dynamics}. Communication latency and jitter are represented by  \(\tau_{ij}(t)\).  \(d_i(t)\) and  \(w_i(t)\) model the exogenous disturbance inputs and stochastic noise components, respectively. In addition, the scheduling delays and protocol-induced errors are embedded within the non-linear structure of the function  \(f_i(\cdot)\). The clock drift and synchronization mismatches are implicitly modeled within the internal state formulation, allowing the model to represent asynchronous interactions and temporal misalignment across the system. 

% This formalism is essential for the design and analysis of:
% \begin{itemize}
%     \item \textbf{Delay-tolerant control systems}, capable of maintaining performance under asynchronous operation.
%     \item \textbf{Secure distributed optimization algorithms}, resilient to data injection and eavesdropping.
%     \item \textbf{Wide-Area Monitoring, Protection, and Control (WAMPAC)} infrastructures, where time synchronization and rapid response are vital.
%     \item \textbf{Coordination of Distributed Energy Resources (DERs)}, involving decentralized protocols and dynamic reconfiguration.
% \end{itemize}

% The practical applications of communication-aware dynamics into control design are:
% \begin{itemize}
%     \item Voltage and frequency regulation using delayed feedback in real time.
%     \item Under communication faults, load shedding and restoration scheduling are possible.
%     \item Synchronization of microgrid and black-start coordination is possible.
%     \item It allows for topology reconfiguration and routing control, responding to cyber and physical disruptions.
% \end{itemize}

% The communication and information dynamics layer forms a foundation to enable real time gird-intelligence. Through a rigorous model, data integrity, signal fidelity and time guarantees are preserved. This helps support integration of AI-based control agents, RL optimizers and robust distributed control frameworks. Altogether, this bridges the modeling gap between physical infrstructure and cyber protocols. 

\subsection{Energy Management System}

EMS is the central cyber-physical component incharge of real-time optimization. It is responsible for coordination of distributed generation, controllable loads and storage units. It receives the data from sensors, it forecasts and then has the ability to execute control logic and communicate the control decisions to the field devices and operators. Altogether, it is the brain of the system striving to achieve system’s operational efficiency, reliability and sustainability. This section elaborates the modeling of the EMS. 

\subsubsection{Optimization Objective}

The first optimization task of the EMS is to ascertain the control inputs  \(u(t)\)  to minimize the overall operational cost of the system in a given time horizon  \([0,T]\). 

\begin{equation}
\min_{u(t)} J = \int_0^T \Bigl(C_G(P_G) + C_L(P_L) + C_E(E) + C_S(u)\Bigr) \, dt.
\end{equation}

Here:
\begin{itemize}
    \item \(C_G(P_G)\): The cost of the generation based on the generation unit’s fuel and ramp rates.
    \item \(C_L(P_L)\): Load Curtailment penalty which essentially reflects customer response. 
    \item \(C_E(E)\): Cost of energy storage. This abstracts degradation of battery lifde and associated dispatch penalties.
    \item \(C_S(u)\): This abstracts cost of control. This is employed to discourage frequent control actions, which induce instability in the system. 
\end{itemize}

The EMS objective combines overall performance over a period, while ensuring multi-objective goals of low operational cost, low emissions, and control optimization are met, keeping the maximization of service reliability as a central piece.

\textbf{Constraints:} The constraints on the objective function described above are grounded in the first-principles involved in the power system. The are given as :

\begin{equation}
\begin{aligned}
&f_{PF}(x,u) = 0,\\
&P_{G_i}^{\min} \leq P_{G_i}(t) \leq P_{G_i}^{\max},\\
&\dot{E}_i(t) = \eta_{ch} P_{ch,i}(t) - \frac{1}{\eta_{dis}} P_{dis,i}(t),\\
&0 \leq P_{shed,i}(t) \leq P_{L_i}(t).
\end{aligned}
\end{equation}

Where:
\begin{itemize}
    \item \(f_{PF}(x,u) = 0\): This is a nonlinear equality constraint. It enforces the physical feasibility of voltage and current variables.
    \item \(P_{G_i}^{\min}\), \(P_{G_i}^{\max}\): System’s Minimum and Maximum capacity of the respective generators. This reflects technical and economic dispatch boundaries.
    \item \(\dot{E}_i(t)\):Reflects the rate of storage for unit. 
    \item \(\eta_{ch}\), \(\eta_{dis}\): Charging and discharging efficiency factors (typically \(0.85 \leq \eta \leq 0.98\)) modeling converter/inverter losses.
    \item \(P_{ch,i}(t)\), \(P_{dis,i}(t)\): At any given time \(t\) , the charging and discharging power of the storage units\(i\).
    \item \(P_{shed,i}(t)\): Load shedding by a given bus \(i\). This is limited to not cross the demand at any given time \(P_{L_i}(t)\).

\end{itemize}

The constraints enable security and sustainability of the operation alongside the integration of the storage, renewables, and felxible demands. 

\subsection{Agent‑Based and Game‑Theoretic Modeling}
\label{sec:Game}
Unlike the traditional grid, modern smart grids are a decentralized and dynamic network of disparate agents, including domestic units, EVs, DER, and energy storage units. These agents interact with the grid bidirectionally and are often called \textit{prosumers}, which produce as well as consume the energy. This highly dynamic behavior can be modeled and optimized through advanced agent-based modeling methodologies and game-theoretic frameworks. These methods present an effective method for distributed decision making suitable for decentralized systems such as the smart grid. 

Utilizing this, we present an agent-based prosumers optimization method in this section. We also present a non-cooperative game-theoretic framework mimicking the strategic interactions between agents for various available resources.

\subsubsection{Prosumer Optimization}

Each prosumer \(i \in \mathcal{P}\) aims to minimize its individual cost function over a planning horizon. The cost function typically accounts for electricity consumption, local generation (e.g., solar PV), battery usage, and potential monetary incentives or penalties. A basic optimization problem for a prosumer is given by:

\begin{equation}
\mathcal{A}_i = \arg\min_{P_i(t)}\left[c_i(P_i(t)) - \lambda(t)P_i(t)\right],
\label{eq:prosumer_opt}
\end{equation}

where:
\begin{itemize}
    \item \(P_i(t)\): Net power exported (positive) or imported (negative) by prosumer \(i\) at time \(t\).
    \item \(c_i(P_i(t))\): Cost or disutility function, typically convex (e.g., quadratic) representing discomfort, battery degradation, or fuel usage.
    \item \(\lambda(t)\): Dynamic locational marginal price (LMP) or incentive rate broadcast by the system operator.
    \item \(\mathcal{A}_i\): The action or strategy space of agent \(i\), such as consumption level, charging schedule, or generation output.
\end{itemize}

The optimization can be subject to local constraints:

\begin{align}
0 &\leq P_i^{\text{load}}(t) \leq P_i^{\text{load,max}}, \\
0 &\leq P_i^{\text{gen}}(t) \leq P_i^{\text{gen,max}}, \\
0 &\leq S_i(t) \leq S_i^{\text{max}},
\end{align}

where \(S_i(t)\) is the state of charge of the local battery.

More sophisticated versions include forecast-based optimization using Model Predictive Control (MPC), where each agent solves:

\begin{equation}
\min_{P_i(t:t+N)} \sum_{k=t}^{t+N} \left[c_i(P_i(k)) - \lambda(k)P_i(k)\right],
\end{equation}

for a prediction horizon \(N\). This supports dynamic decision-making under uncertainty in demand or PV output.

\subsubsection{Non‑Cooperative Game}

In environments where multiple prosumers act simultaneously and selfishly, their actions impact each other through shared variables like voltage, frequency, or market prices. This leads to a non-cooperative game setup:

\begin{equation}
\max_{\mathcal A_i} U_i(\mathcal A_i,\mathcal A_{-i}), \quad \forall i\in\mathcal{N},
\label{eq:game_util}
\end{equation}

where:
\begin{itemize}
    \item \(U_i(\cdot)\): Utility function of agent \(i\).
    \item \(\mathcal{A}_i\): Strategy of agent \(i\).
    \item \(\mathcal{A}_{-i}\): Strategies of all other agents (\(j \neq i\)).
    \item \(\mathcal{N}\): Set of all participating agents or prosumers.
\end{itemize}

A classical solution concept is the \textbf{Nash Equilibrium}, where no agent can unilaterally improve its utility:

\[
U_i(\mathcal{A}_i^*, \mathcal{A}_{-i}^*) \geq U_i(\mathcal{A}_i, \mathcal{A}_{-i}^*), \quad \forall \mathcal{A}_i \in \mathcal{A}_i, \forall i.
\]

Utility functions \(U_i\) can incorporate:
\begin{itemize}
    \item Energy profit: \(\lambda(t)P_i(t)\),
    \item Battery wear cost: \(\beta \cdot \text{DoD}_i(t)\),
    \item Comfort or preference: deviation from desired HVAC or EV charging profile,
    \item Peer-to-peer (P2P) trade benefits or penalties.
\end{itemize}

The existence of a Nash equilibrium is guaranteed under certain conditions such as convexity of utility functions and compactness of strategy sets. For uniqueness, strict convexity or contraction mappings may be needed. Game-theoretic analysis often leads to distributed algorithms, e.g., best-response dynamics or fictitious play.

\subsection{Security and Resilience}
\label{sec:FDI}

As the IoT devices and communication infrastructure is integrated in smart grids, it becomes imperative to ensure strong cybersecurity measures. This section explains the False Data Injections and assessment of system resilience. 

\subsubsection{False Data Injection}

One of the most prominent and dangerous classes of cyberattacks in smart grids is the False Data Injection (FDI) attack. FDI attacks aim to compromise the integrity of sensor measurements by inserting malicious data that can mislead state estimation, control actions, or optimization routines without being detected by standard bad-data detection schemes. Mathematically, if \( y_i(t) \) is the true sensor measurement at time \( t \), then the compromised measurement under an FDI attack is modeled as:

\begin{equation}
\tilde{y}_i(t) = y_i(t) + a_i(t),
\end{equation}

where \( a_i(t) \) denotes the adversarial signal injected into the measurement. In a coordinated attack, the adversary may exploit the topology of the communication graph and knowledge of the system model to design \( a_i(t) \) such that the corrupted measurement passes standard residual checks used in state estimation algorithms. Advanced techniques for FDI detection include machine learning-based anomaly detection, robust state estimation, and statistical change detection. However, securing all communication nodes and sensors in large-scale smart grids remains a nontrivial challenge due to cost and scalability concerns.

\subsubsection{Resilience Index}

System resilience defines the grid’s ability to perceive, absorb, adapt and recover from disruptions. The disruptions may be cyberattacks but also physical faults and natural disasters such as wildfires. Resilience Index(RI) is the metric which comparatively quantifies the system’s deviation from its nominal behavior during a disturbances. 

The resilience index used for evaluation is
\begin{equation}
R = 1 -
\frac{\sum_k \| x_k - x_k^{\mathrm{nom}} \|^2}
     {\sum_k \| x_k^{\mathrm{nom}} \|^2},
\end{equation}
where $x_k^{\mathrm{nom}}$ is the nominal state trajectory. This metric, computed directly in the simulation, quantifies the ability of the controller to maintain stable, nominal-like operation during disturbances and cyberattacks. Factors affecting the resilience include deployment of redundant and diverse control mechanisms, EMS responsiveness to anomalous behavior, the strength of the communication network, and the autonomous ability of AI-based adaptive controllers to reconfigure operations.

\section{Adaptive Control Methodology}
\label{sec:control}
The adaptive control layer governs the control response to external disturbance and prosumers' behavior. The adaptive control is scalable and robust due to its ability to learn directly from data and not the models.Two foundations in our approach are adaptive dynamic programming(ADP), Proximal Policy Optimization (PPO) and Deep Q-Network (DQN) for long-term performance.
\begin{figure}[h!]
    \centering
    \includegraphics[width=0.95\linewidth]{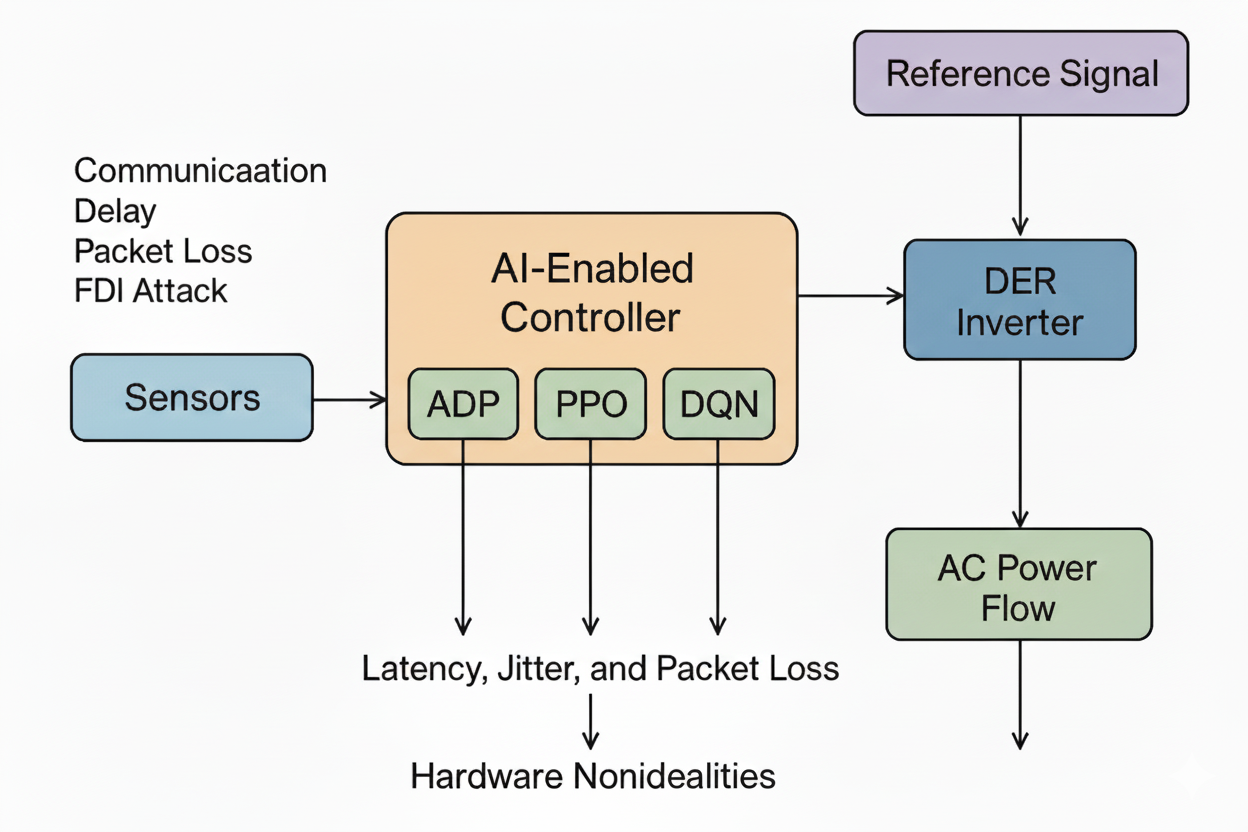}
    \caption{Hybrid AI-enabled cyber-physical control architecture illustrating sensing, communication delay, packet loss, AI controllers (ADP, PPO, DQN), inverter actuation, and AC power flow dynamics.}
    \label{fig:control}
\end{figure}

\subsection{Adaptive Dynamic Programming}

Adaptive Dynamic Programming (ADP) provides a framework to approximate solutions to complex dynamic optimization problems using learning-based techniques. It extends Bellman's principle of optimality to problems with unknown or nonlinear dynamics.

The core of ADP involves minimizing the expected cumulative cost-to-go. For a state \(x_t\), control \(u_t\), and cost function \(r(x,u)\), the optimal control \(u^*(t)\) satisfies:
\begin{equation}
u^*(t) = \arg\min_u \left[r(x_t,u_t) + \gamma \hat{V}(x_{t+1})\right],
\label{eq:adp_policy}
\end{equation}
where \(r(x_t,u_t)\) is the immediate cost incurred at time \(t\), \(\hat{V}(x_{t+1})\) is the learned value function that approximates the expected future cost starting from state \(x_{t+1}\), and \(\gamma \in (0,1]\) is the discount factor that reflects the relative importance of future rewards. The learned value function \(\hat{V}(x)\) serves as an approximation to the true optimal cost-to-go function \(V^*(x)\), which is defined as:
\begin{equation}
V^*(x_t) = \min_{\{u_k\}_{k=t}^\infty} \sum_{k=t}^\infty \gamma^{k-t} r(x_k, u_k),
\label{eq:optimal_value}
\end{equation}
subject to the system dynamics \(x_{k+1} = f(x_k, u_k)\). Since the exact function \(V^*(x)\) is generally unknown or intractable in high-dimensional, nonlinear systems, ADP approximates it using a parameterized function:
\begin{equation}
\hat{V}(x; \theta) = \sum_{i=1}^N \theta_i \phi_i(x) = \Phi(x)^\top \theta,
\label{eq:value_function}
\end{equation}
where \(\phi_i(x)\) are chosen basis functions (e.g., radial basis functions, polynomials, or neural network activations), \(\theta_i\) are trainable weights, and \(\Phi(x)\) is the feature vector. Alternatively, \(\hat{V}(x)\) may be implemented using a neural network denoted by \(\mathcal{N}_\theta(x)\), where \(\theta\) represents the learnable parameters of the network.

The immediate cost function \(r(x_k, u_k)\) quantifies the penalty associated with the current state and control input. It is typically designed to penalize deviations from desired operating conditions and to discourage excessive control effort. A commonly used quadratic form is:
\begin{equation}
r(x_k, u_k) = (x_k - x_{\text{ref}})^\top Q (x_k - x_{\text{ref}}) + u_k^\top R u_k,
\label{eq:immediate_cost}
\end{equation}
where \(x_{\text{ref}}\) is the reference state (e.g., nominal frequency or voltage), \(Q \succeq 0\) is a state weighting matrix, and \(R \succ 0\) is a control weighting matrix. This cost function penalizes deviations from the states as well as the control actions so it ensures stable and efficient operation. A simple form of the cost function has the form of:
\begin{equation}
r(x_k, u_k) = \omega_k^2 + \alpha u_k^2,
\end{equation}
where \(\omega_k\) is the frequency deviation and \(u_k\) is the control input from the governor or AVR, and \(\alpha > 0\) controls the trade-off between tracking performance and control effort. By iteratively updating both the value function approximation \(\hat{V}(x)\) and the optimal control \(u^*(t)\), ADP enables near-optimal decision-making in real time, even in the presence of system uncertainties and nonlinear dynamics. In smart grids, this framework allows agents (e.g., DER controllers, load-serving entities) to update control actions in real-time based on observed system evolution. Unlike traditional dynamic programming, ADP handles high-dimensional state-action spaces via function approximators (e.g., neural networks).

\subsection{Proximal Policy Optimization (PPO) Control Methodology}

Proximal Policy Optimization (PPO) is a policy-gradient reinforcement learning algorithm that enables stable and sample-efficient control in nonlinear dynamic systems. It is particularly suitable for smart grid applications where the control space is continuous (e.g., inverter reactive power modulation, EV charging rates) and where measurement disturbances, cyber-latency, and stochastic behavior of loads require robustness. PPO optimizes a parameterized control policy $\pi_\theta(u|x)$ that maps system states to control actions, enabling fast real-time decision-making in decentralized energy resources.

The goal of PPO is to maximize the expected return:
\begin{equation}
J(\theta) = \mathbb{E}_{\pi_\theta} \left[ \sum_{k=t}^{\infty} \gamma^{k-t} \, r(x_k, u_k) \right],
\label{eq:ppo_objective}
\end{equation}
where $\gamma \in (0,1]$ is the discount factor and $r(x_k,u_k)$ is the instantaneous cost or reward derived from system performance. In the context of frequency and voltage regulation in a grid, the reward typically penalizes frequency deviations, voltage violations, and excessive inverter effort. Unlike classical policy gradient methods, PPO constrains policy updates using a clipped surrogate objective designed to prevent large detrimental gradient steps. The clipped objective for PPO is:
\begin{equation}
L^{\text{clip}}(\theta) =
\mathbb{E}_t \left[
\min \left(
r_t(\theta) \, \hat{A}_t, \;
\text{clip}\big(r_t(\theta), 1 - \epsilon, 1 + \epsilon\big) \hat{A}_t
\right)
\right],
\label{eq:ppo_clip}
\end{equation}
where 
\[
r_t(\theta) = 
\frac{\pi_\theta(u_t|x_t)}{\pi_{\theta_{\text{old}}}(u_t|x_t)}
\]
is the likelihood ratio and $\hat{A}_t$ is the advantage estimate computed using generalized advantage estimation (GAE). The clipping threshold $\epsilon$ limits how far the new policy can deviate from the old one at each update, significantly improving stability under noisy smart-grid measurements.

The agent uses a parameterized policy network (actor) to output continuous control actions:
\[
u_t = \pi_\theta(x_t),
\]
and a value network (critic) to approximate the state-value function:
\[
V_\psi(x_t) \approx \hat{V}(x_t) =
\mathbb{E}\left[ \sum_{k=t}^{\infty} \gamma^{k-t} r(x_k, u_k) \right].
\]

In smart grid applications, the state vector includes bus voltages, angles, renewable power availability, SOC of batteries, EV charging levels, and cyber-layer indicators such as delay and packet loss. PPO updates both the actor and critic networks with batched trajectory data from the cyber -- physical simulation environment. This leads to a robust controller that is able to compensate the renewable intermittency and measure corruption and stabilize system frequency.

To specialize PPO to perform grid control, the reward function is defined as follows:
\begin{equation}
r(x_k, u_k) = 
- \left(
\omega_k^2 + 
\beta_v (V_k - 1)^2 + 
\alpha \| u_k \|^2
\right),
\label{eq:ppo_reward}
\end{equation}
where $\omega_k$ is the frequency deviation, $V_k$ is the voltage magnitude and $\alpha, \beta_v > 0$ are the penalties for excessive control effort and voltage deviations. This reward structure enables PPO to learn smooth and stable, safe control actions even in the presence of network delay, cyber-attacks, and hardware uncertainties.

In the proposed hybrid control framework, PPO is the high-level continuous-action controller that is in charge of real-time tuning of inverter outputs, battery charge/discharge levels and flexible load-tuning. To noisy gradients and its capacity to manage huge nonlinear state spaces make it suitable for next generation adaptive smart grid operations.

\subsection{Deep Q-Network (DQN) Control Methodology}

Deep Q-Networks (DQN) is a value-based reinforcement learning algorithm that is applicable for discrete control problems in cyber--physical smart grids. While PPO is used for continuous control signals, DQN is used for discrete decision-making like mode switching (charging/discharging states of storage,) EV charging priority assignment, load-shedding triggers and inverter operation modes. The method is an extension to classical Q-learning in which Q(x, u) is approximated by a deep neural network which allows efficient learning in large nonlinear state spaces.

DQN aims to approximate the optimal $Q$-function which is given by the Bellman optimality equation:mate the optimal $Q$-function defined by the Bellman optimality equation:
\begin{equation}
Q^*(x_t,u_t) = 
r(x_t,u_t) + \gamma \max_{u'} Q^*(x_{t+1},u'),
\label{eq:dqn_bellman}
\end{equation}
which represents the maximum future return achievable from state $x_t$ by taking action $u_t$. Since directly computing $Q^*$ is intractable for high-dimensional systems like smart grids, DQN uses a parameterized function $Q_\theta(x,u)$ to approximate it:
\[
Q_\theta(x_t, u_t) \approx Q^*(x_t,u_t),
\]
where $\theta$ denotes the neural network parameters. The network parameters are updated by minimizing the temporal-difference loss:
\begin{equation}
\mathcal{L}(\theta) = 
\left( r_t + \gamma \max_{u'} Q_{\theta^-}(x_{t+1},u') - Q_\theta(x_t,u_t) \right)^2,
\label{eq:dqn_loss}
\end{equation}
where $Q_{\theta^-}$ is the target network, updated slowly to improve training stability. Experience replay buffers further decorrelate training samples, preventing divergence and enabling stable learning even when the environment is highly nonlinear and stochastic. In the context of the proposed smart grid architecture, the state $x_t$ contains voltage profiles, renewable forecasts, load levels, SOC of batteries, and cyber-health indicators (latency, packet success rate). The discrete action set includes:
\begin{itemize}
    \item battery mode selection (charge/discharge/idle),
    \item EV charging priority states,
    \item inverter tap or mode changes,
    \item load-shedding or restoration triggers.
\end{itemize}

To tailor the DQN controller for grid stability, the reward function is defined as:
\begin{equation}
r(x_k,u_k) =
- \left(
\omega_k^2 + \lambda \, \mathbb{I}\{\text{voltage violation}\} 
+ \alpha \|u_k\|^2
\right),
\label{eq:dqn_reward}
\end{equation}
where $\mathbb{I}\{\cdot\}$ denotes an indicator function and $\lambda$ penalizes actions that result in unacceptable voltage profiles. This reward formulation encourages DQN to select discrete operational modes that indirectly support frequency and voltage stability, especially during fast renewable ramps or cyber-induced disturbances. DQN is thus integrated into the hybrid control framework as a complementary discrete-action controller operating alongside the continuous-action PPO controller. Together with the ADP layer, DQN contributes to hierarchical decision making where discrete grid management actions are coordinated with continuous inverter and storage control signals, creating a unified adaptive control mechanism for complex cyber--physical smart grids.

% *************************************************************
% ===============================================================
% \section{Cyber‑Physical Power System Testbed}

\section{Modeling and implementation of Testbed}
\label{sec:System}
% ===============================================================

The physical power system, measurement infrastructure, and cyber-physical interaction architecture on which the proposed control methodology is implemented emulated in the simulation environment which consist of a physical layer i.e. modified IEEE 33-bus radial distribution system, enriched with distributed energy resources (DERs), electric vehicles (EVs), responsive loads, and supervisory communication layers. The physical layer is tightly coupled with the cyber layer, enabling realistic evaluation under communication delay, packet drops, measurement corruption, and cyberattacks.

%  A modified IEEE 33-bus radial distribution system, a standard benchmark for resilience and control studies. The feeder consists of:
% \begin{itemize}
%     \item 33 buses and 32 distribution lines,
%     \item nominal voltage of 12.66~kV (line-to-line),
%     \item total active load of approximately 3.72~MW,
%     \item total reactive load of 2.3~Mvar,
%     \item four laterals with varying load densities,
%     \item integration of renewable and controllable DER units.
% \end{itemize}

The network is modeled using a nonlinear differential--algebraic system:
\begin{align}
\dot{x} &= f(x,y,u), \label{eq:phys_dyn1} \\
z &= g(x,y),            \label{eq:phys_dyn2}
\end{align}
where $x$ represents dynamic states (e.g., inverter states, SOC), $y$ contains algebraic variables (voltages, angles), and $u$ represents the control actions synthesized by the hybrid RL-based controller. The physical power flows are computed using full nonlinear AC equations, ensuring realism in voltage, frequency, and power exchange behavior.

% ----------------------------------------------------------
\subsection{Loads and Demand Profiles}
% ----------------------------------------------------------

The loads at each bus are modeled as time-varying active and reactive power demands:
\begin{align}
P_{L,k} &= P_{L}^{\mathrm{base}} (1 + \Delta P_k), \\
Q_{L,k} &= Q_{L}^{\mathrm{base}} (1 + \Delta Q_k),
\end{align}
where the perturbations $\Delta P_k$ and $\Delta Q_k$ follow stochastic time-series patterns emulating residential, commercial, and EV-driven peak variations. Sudden changes of up to 20\% are introduced to emulate high-variability events, forcing the controller to react in real time. An aggregated EV charging block is modeled as an adjustable active power load $E^{EV}_k$, affecting both voltage stability and system frequency during high charging demand.

% ----------------------------------------------------------
\subsection{Renewable Generation and DER Modeling}
% ----------------------------------------------------------

The network includes solar PV and wind DER units placed at selected buses. Their injections follow:
\begin{align}
S_k^{\mathrm{solar}} &= P_{\mathrm{PV}}(k) + jQ_{\mathrm{PV}}(k), \\
S_k^{\mathrm{wind}}  &= P_{\mathrm{W}}(k) + jQ_{\mathrm{W}}(k),
\end{align}
with their active power governed by real irradiance and wind-speed profiles:
\begin{align}
P_{\mathrm{PV}}(k) &= \eta_{\mathrm{PV}} A_{\mathrm{PV}} I_{\mathrm{solar}}(k), \\
P_{\mathrm{W}}(k)  &= \tfrac{1}{2} \rho A C_p v_{\mathrm{wind}}^3(k).
\end{align}

To emulate renewable intermittency, abrupt ramps and cloud-induced drops of 20--40\% are introduced. These disturbances strongly impact bus voltages and frequency, making them ideal for testing adaptive RL controllers.

% ----------------------------------------------------------
\subsection{Battery Storage and EV Modeling}
% ----------------------------------------------------------

Battery storage units are included with the state-of-charge (SOC) dynamic:
\begin{equation}
SOC_{k+1} = SOC_k + \eta_c P^{\mathrm{ch}}_k - \frac{1}{\eta_d} P^{\mathrm{dis}}_k,
\end{equation}
where $P^{\mathrm{ch}}$ and $P^{\mathrm{dis}}$ represent charging/discharging commands derived from the selected controller. SOC affects both frequency regulation and voltage stability through local balancing. EVs are modeled as controlled loads whose demand $E^{EV}_k$ can be modulated by the RL agents to maintain grid stability during congested periods.

% ----------------------------------------------------------
\subsection{Measurement Layer and State Acquisition}
% ----------------------------------------------------------

Each bus is assumed to have a PMU-like sensing layer capable of reporting:
\begin{equation}
s_k = \left[V_k,\;\theta_k,\;P_{L,k},\;Q_{L,k},\;S_k^{\mathrm{solar}},\;
S_k^{\mathrm{wind}},\;SOC_k,\;E^{EV}_k,\;f_k\right]^{\top}.
\end{equation}

These measurements form the state vector used by all controllers. Sensor update rates are 1 second, matching real-time distribution-level monitoring systems. Measurement noise is modeled as:
\begin{equation}
v_k \sim \mathcal{N}(0,\sigma_v^2),
\end{equation}
with $\sigma_v^2$ chosen to represent typical PMU and micro-PMU noise characteristics.

% ----------------------------------------------------------
\subsection{Cyber Layer, Communication Model, and Attacks}
% ----------------------------------------------------------

The communication layer is modeled using a time-varying latency variable $\tau_k$ (0--250 ms), a packet-loss probability $p_{\mathrm{drop}}$, and an FDI attack vector $a_k$:
\begin{equation}
\tilde{s}_k =
\begin{cases}
s_{k-\tau_k} + v_k + a_k, &\text{with probability } 1-p_{\mathrm{drop}},\\
s_{k-1}, &\text{if a packet is dropped}.
\end{cases}
\end{equation}

The FDI attacks manipulate voltage magnitude and frequency to mislead controllers. Time-synchronized bursts occur randomly to emulate coordinated cyberattacks. This cyber-physical coupling makes it necessary for the controller to constantly cope with delayed or outdated states, with corrupted measurements, with missing communications, and even on instability caused by an attack.

% ----------------------------------------------------------
\subsection{Actuation and Inverter Control Model}
% ----------------------------------------------------------

Controller outputs affect the physical network through inverter-based DERs and controllable loads. The applied action includes edge noise:
\begin{equation}
u_k^{\mathrm{applied}} = u_k + e_k, \qquad 
e_k \sim \mathcal{N}(0,\sigma_{\mathrm{edge}}^2),
\end{equation}
representing inverter switching nonidealities and real hardware uncertainty. 
Typical actions of inverters are reactive power injection or absorption, active power curtailment, voltage reference modification and frequency support. These actions influence the subsequent AC power-flow solution and thus the next state $s_{k+1}$. The full system is comprised of a nonlinear radial distribution grid and time-varying loads and renewable energy sources, battery and EV dynamics, PMU-like sensing devices, communication delays and noise, cyberattacks in the form of false data injection (FDI), and various actuator noise sources and inverter nonidealities.

This tightly integrated cyber--physical structure provides a realistic platform for deploying the hybrid ADP--PPO--DQN control methodology described in Section~\ref{sec:Control}. The system reflects the operational conditions of next-generation smart grids, where both physical disturbances and cyber vulnerabilities must be simultaneously addressed.

% **************************************************************

% ============================================
\section{Practical Prototypic System Specifications}
\label{sec:practical-system}
% ============================================

To validate the hybrid cyber--physical control methodology described in Section~\ref{sec:control},
a detailed prototypic distribution-level power system was constructed in the simulation
environment. The system emulates a medium-voltage radial feeder equipped with
distributed energy resources (DERs), electric vehicles (EVs), and advanced metering
infrastructure (AMI). The numerical values of system parameters, communication
characteristics, and physical component models were selected to reflect realistic
operating conditions typically observed in practical distribution networks such as the IEEE~33-bus.

The following subsections summarize the physical and cyber parameters used for the
implementation of the proposed control methodology.

% ----------------------------------------------
\subsection{Network Topology and Electrical Parameters}
% ----------------------------------------------

The prototypic grid consists of a radial feeder with 33 buses, a base voltage of \(12.66~\mathrm{kV}\), a base power of \(10~\mathrm{MVA}\), and a nominal frequency of \(50~\mathrm{Hz}\). The feeder contains a mix of residential, commercial, and small industrial loads with time-varying characteristics. Transformer parameters are modeled with a resistance of \(R_\text{tr} = 0.01~\mathrm{p.u.}\), a reactance of \(X_\text{tr} = 0.04~\mathrm{p.u.}\), and a load-to-no-load ratio of \(T_\text{load}/T_\text{no-load} = 0.8/0.2\). For line segments, typical underground or overhead conductor impedances are assumed, with resistance \(R_\ell \in [0.03,\,0.09]~\Omega/\text{km}\) and reactance \(X_\ell \in [0.04,\,0.12]~\Omega/\text{km}\), and segment lengths ranging from 0.6~km to 1.9~km depending on the bus number. Nodal voltage limits are enforced according to standard utility practice, with \(0.95 \leq V_k \leq 1.05~\mathrm{p.u.}\).

% ----------------------------------------------
\subsection{Load, Renewable, and Storage Modeling}
% ----------------------------------------------

The physical system incorporates time-varying loads, stochastic renewable generation, and distributed storage units to emulate real-world operational variability. The load profiles include active loads \(P_{L,k} \in [0.2,\,1.8]~\text{MW}\), reactive loads \(Q_{L,k} \in [0.05,\,0.9]~\text{MVAr}\), daily variability of 15--30\%, and stochastic fluctuations modeled as Gaussian noise with \(\sigma_{P}=0.03~\text{MW}\). The solar PV system has a rated power of \(S_k^{\rm solar,max} = 500~\text{kW}\), with intermittency modeled via Beta-distributed irradiance and cloud-induced dips ranging from 20--60\% of nominal output. The wind generator features a rated power of \(S_k^{\rm wind,max} = 300~\text{kW}\), a wind speed following a Weibull distribution with scale \(c=7.5\), and cut-in/cut-out limits between 3--22~m/s. The battery energy storage system (BESS) is modeled with a maximum energy capacity of \(E_{\rm max} = 300~\text{kWh}\), maximum charge/discharge power \(P_{\rm max} = \pm 150~\text{kW}\), and state-of-charge limits \(SOC_k \in [0.20,\,0.90]\), with a charging/discharging efficiency of 95\%. The electric vehicle charging load ranges from \(E^{\rm EV}_k \in [40,\,110]~\text{kW}\) with an efficiency of \(\eta_\text{EV} = 0.92\).

% ----------------------------------------------
\subsection{Cyber Layer, Network Delay, and Attack Parameters}
% ----------------------------------------------

The communication layer emulates practical conditions in AMI and edge-to-cloud coordination, where each state vector reported by bus controllers is affected by latency, packet drops, measurement noise, and false data injection (FDI) attacks. The communication delay is modeled as a discrete-time variable \(\tau_k \in \{0,1,2,3\}~\text{steps}\) (equivalent to 20--120~ms), following a truncated Gaussian distribution. Packet loss is represented by a probability \(p_{\rm drop} = 0.05\), reflecting congestion and wireless interference typical in edge networks. Measurement noise is included for voltage and frequency, with \(v_{V,k} \sim \mathcal{N}(0,\,0.005^2)\) and \(v_{f,k} \sim \mathcal{N}(0,\,0.02^2)\), consistent with real PMU/µPMU accuracy. FDI attacks modify the voltage and frequency as \(a_{V,k} \in [-0.03,\,0.03]~\mathrm{p.u.}\) and \(a_{f,k} \in [-0.15,\,0.18]~\mathrm{Hz}\), applied with a probability \(p_{\rm FDI} = 0.04\) during attack intervals. Control signals sent to DER inverters experience physical uncertainty:
\begin{align}
    u_{k}^{\mathrm{applied}} = u_{k} + e_{k}, \qquad
    e_k \sim \mathcal{N}(0,\,0.01^2).
\end{align}

Inverter response dynamics follow a first-order model:
\begin{equation}
    \tau_{\rm inv} \dot{P} + P = P_{\rm ref},\qquad 
    \tau_{\rm inv} = 40~\text{ms}.
\end{equation}

% % ----------------------------------------------
% \subsection{Simulation Discretization and Time Horizon}
% % ----------------------------------------------

% The dynamic equations of the grid are solved using a forward Euler method with:
% \begin{align}
%     \Delta t &= 20~\text{ms},\\
%     T_{\rm sim} &= 6000~\text{steps} \approx 120~\text{s}.
% \end{align}

% This short-duration window is representative of fast-timescale voltage and frequency control in distribution grids.

% % ----------------------------------------------
% \subsection{Summary of Practical System Characteristics}
% % ----------------------------------------------

% The prototypic system integrates realistic network impedances, renewable intermittency,
% EV charging variability, BESS constraints, stochastic loads, communication delay,
% packet loss, and FDI cyberattacks. All parameters are grounded in ranges typically
% observed in utility-grade distribution systems. This ensures that the evaluation of
% the proposed hybrid ADP–PPO–DQN control framework is representative of practical
% medium-voltage grid behavior under real-world cyber and physical uncertainties.

% **************************************************************

\section{Implementation of Control Methodology}
\label{sec:Control} 
This section presents the mathematical formulation of the hybrid cyber--physical control architecture implemented in the simulation engine. The controller integrates Adaptive Dynamic Programming (ADP), Proximal Policy Optimization (PPO), and Deep Q-Networks (DQN) under communication constraints, packet drops, measurement corruption, and false-data injection (FDI) cyberattacks. 

\subsection{State Vector and Cyber Observation Model} 
At each discrete control step $k$, each bus provides a measurement vector
\begin{equation}
s_k = \big[\,V_k,\theta_k,P_{L,k},Q_{L,k},S^{\mathrm{solar}}_k,
S^{\mathrm{wind}}_k,SOC_k,E^{EV}_k,f_k\,\big]^{\top}
\end{equation}
where $V_k$ is voltage (p.u.), $\theta_k$ is angle (rad), $P_{L,k}$ and $Q_{L,k}$ are active/reactive loads, $S^{\mathrm{solar}}_k$ and $S^{\mathrm{wind}}_k$ are renewable injections, $SOC_k$ is battery state-of-charge, $E^{EV}_k$ is EV charging load, and $f_k$ is frequency (Hz). These correspond directly to the features normalized in the simulation. The cyber layer distorts the measurement due to delay, packet loss, noise, and FDI attacks. Let $\tau_k$ be the communication delay, $p_{\mathrm{drop}}$ the packet loss probability, $v_k$ the measurement noise, and $a_k$ the FDI corruption. The controller receives
\begin{equation}
\tilde{s}_k =
\begin{cases}
s_{k-\tau_k} + a_k + v_k, & \text{with probability } 1-p_{\mathrm{drop}},\\[3pt]
s_{k-1}, & \text{if a packet drop occurs.}
\end{cases}
\end{equation}
here delays are discretized, FDI modifies voltage and frequency, and packet loss forces fallback to the previous state.

\subsection{Unified Control Objective}

All controllers minimize the same quadratic cost
\begin{equation}
c_k = (V_k - 1)^2 + (f_k - 50)^2 + \alpha \sum_{i=1}^m u_{k,i}^2,
\label{eq:cost}
\end{equation}
which penalizes deviations from nominal voltage and frequency and includes a control-effort penalty weighted by $\alpha>0$. The reward used in reinforcement learning is
\begin{equation}
r_k = -c_k.
\end{equation}

\subsection{Adaptive Dynamic Programming (ADP)}
ADP approximates the value function
\begin{equation}
V(s_k) \approx \hat{V}(s_k;\phi),
\end{equation}
where $\phi$ denotes neural network parameters. The policy network produces a continuous control action
\begin{equation}
u_k^{\mathrm{ADP}} = \pi(s_k;\theta),
\end{equation}
with parameters $\theta$. The temporal-difference (TD) target and error are
\begin{equation}
y_k = r_k + \gamma \hat{V}(s_{k+1};\phi), \qquad
\delta_k = y_k - \hat{V}(s_k;\phi).
\end{equation}
Both edge and cloud ADP controllers are implemented:
\begin{align}
u_k^{\mathrm{edge}} &= \pi(\tilde{s}_k),\\
u_k^{\mathrm{cloud}} &= \pi(s_k),
\end{align}
where the cloud version benefits from cleaner states but suffers communication delay.

\subsection{Proximal Policy Optimization (PPO)}

PPO uses a Gaussian stochastic policy
\begin{equation}
u_k^{\mathrm{PPO}} \sim \mathcal{N}\!\left(\mu_\theta(\tilde{s}_k),\, \sigma_\theta^2(\tilde{s}_k)\right),
\end{equation}
where $\mu_\theta$ and $\sigma_\theta$ are outputs of the actor network. The importance ratio is
\begin{equation}
\rho_k(\theta) = 
\frac{\pi_\theta(a_k \mid s_k)}{\pi_{\theta_{\mathrm{old}}}(a_k \mid s_k)},
\end{equation}
and the clipped PPO objective is
\begin{equation}
L^{\mathrm{PPO}}(\theta) =
\mathbb{E}\!\left[
\min\!\left(
\rho_k(\theta) A_k,\;
\mathrm{clip}\!\left(\rho_k(\theta),1-\epsilon,1+\epsilon\right) A_k
\right)
\right],
\end{equation}
with advantage estimate $A_k = r_k - V(s_k)$. Packet loss is implemented as
\begin{equation}
u_k^{\mathrm{PPO,applied}} =
\begin{cases}
u_k^{\mathrm{PPO}}, & \text{with probability } 1-p_{\mathrm{drop}},\\
0, & \text{if packet dropped}.
\end{cases}
\end{equation}

\subsection{Deep Q-Network (DQN)}

The continuous action range is discretized into a finite set $\mathcal{A}$. The Q-network approximates
\begin{equation}
Q(s_k,a;\theta_q).
\end{equation}
Action selection is $\epsilon$-greedy:
\begin{equation}
a_k^{\mathrm{DQN}} =
\begin{cases}
\arg\max_{a \in \mathcal{A}} Q(\tilde{s}_k,a), & \text{with prob. } 1-\epsilon,\\
\text{random element of }\mathcal{A}, & \text{with prob. } \epsilon.
\end{cases}
\end{equation}
The TD update is
\begin{equation}
Q_{\theta_q}(s_k,a_k) \leftarrow
r_k + \gamma \max_{a'} Q_{\theta_q}(s_{k+1},a').
\end{equation}

\subsection{Actuation Model and Cyber Noise}

The physically applied action includes control noise representing edge uncertainty:
\begin{equation}
u_k^{\mathrm{applied}} = u_k + e_k, \qquad
e_k \sim \mathcal{N}(0,\sigma_{\mathrm{edge}}^2),
\end{equation} 
The grid dynamics follow the differential–algebraic model
\begin{align}
\dot{x} &= f(x,y,u_k^{\mathrm{applied}}), \\
0 &= g(x,y),
\end{align}
which determines the next physical state $s_{k+1}$.

\subsection{Hybrid Controller Coordination}

At each bus and time step, the supervisory layer evaluates the instantaneous cost of each controller:
\begin{align}
c_k^{\mathrm{ADP}},\qquad
c_k^{\mathrm{PPO}},\qquad
c_k^{\mathrm{DQN}},
\end{align}
computed using \eqref{eq:cost}. The hybrid supervisor selects
\begin{equation}
u_k^{*} = \arg\min_{u \in \{u^{\mathrm{ADP}},u^{\mathrm{PPO}},u^{\mathrm{DQN}}\}}
c_k(u).
\end{equation}
\subsection{Algorithm}
The proposed algorithm of AI Enhanced Hybrid Cyber-Physical Adaptive Control incorporates real time sensing, state estimation and adaptive decision making to optimise the performance of smart grid. It makes use of neural network based policies and Approximate Dynamic Programming (ADP) to iteratively generate and evaluate the control actions. At each time step, measurements from PMUs, SCADA and IoT sensors are processed to estimate the system state and candidate control signals are then generated and evaluated with respect to a value function. Learning is done through temporal-difference updates and reinforcement feedback so that the policy performance is continuously improved. 
\begin{algorithm}[h!]
\caption{AI-Enhanced Hybrid Cyber‑Physical Adaptive Control for Smart Grid}
\KwData{\\
System graph $\mathcal{G} = (\mathcal{N}, \mathcal{E})$; \\
Dynamic states $x(t)$; \\
Measurements $y(t)$ from SCADA/PMUs; \\
Control objectives $\mathcal{J}$ (e.g., cost minimization, stability); \\
Neural Network policy model $\pi_{\theta}(x)$; \\
Value function approximator $\hat{V}_{\phi}(x)$
}
\KwResult{Intelligent control signals $u^*(t)$ to stabilize and optimize grid performance}

\textbf{Initialization:} \\
Define physical dynamics $\dot{x}(t) = f(x,u)$\; 
Define cyber measurements $y = h(x)$\; 
Initialize neural network weights $\theta$, $\phi$\;

\While{grid is operational}{
    \textbf{Sensing:} Acquire $y(t)$ from PMUs, SCADA, and IoT sensors\;

    \textbf{State Estimation:} Estimate $\hat{x}(t)$ using Kalman filter or AI-based state predictor\;

    \textbf{AI-Based Control Policy Update:} \\
    Use neural network policy $\pi_{\theta}(x)$ to generate candidate control actions\;

    \textbf{ADP-Based Evaluation:} \\
    Compute optimal action using: \\
    $u^*(t) = \arg\min_u \left[r(x,u) + \gamma \hat{V}_{\phi}(x_{t+1})\right]$\;

    \textbf{Learning:} \\
    Update $\hat{V}_{\phi}(x)$ using temporal difference learning\; 
    Update policy $\pi_{\theta}(x)$ using reinforcement feedback (e.g., policy gradient, actor-critic)\;

    \textbf{Execution:} \\
    Apply $u^*(t)$ to grid actuators (generators, FACTS, inverters)\;

    \textbf{Resilience Evaluation:} \\
    Simulate perturbations or faults and adapt policy accordingly\;
}
\Return{Stable and intelligent dispatch trajectory across cyber‑physical system}
\end{algorithm}

% ===============================================================

\begin{table*}[h!]
\centering
\small
\caption{Data of System Measurements and Control for the timestamp value 11/1/2025 00:00}
\makebox[\textwidth][c]{%
\begin{adjustbox}{max width=\textwidth}
\begin{tabular}{cccccccccccccccccccccccccccc}
\toprule
Timestamp & BusID & Voltage(pu) & Angle(rad) & LoadP(kW) & LoadQ(kVAR) & SolarGen(kW) & WindGen(kW) & BatterySOC(\%) & EVLoad(kW) & Frequency(Hz) & FeederPower(kW) & CommunicationDelay(ms) & PacketLossRate & FDI\_Attack & FDI\_Severity & MeasurementError & ADP\_ValueFunction & ADP\_ControlAction & PPO\_Action & PPO\_Reward & DQN\_QValue & DQN\_Action & ResilienceIndex & Edge\_EstimatedState & Cloud\_OptimizedState & Edge\_ControlAction & Cloud\_UpdateFlag \\
\midrule

11/1/2025 0:00 & 1 & 0.954 & 0.00253 & 120.919 & 35.085 & 0 & 4.627 & 49.411 & 0 & 50.0413 & 139.134 & 46.861 & 0.02818 & 0 & 0 & 0.006425 & 0.4666 & 0.3831 & 2 & -0.03769 & 0.15631 & 0 & 0.93896 & V=0.954;A=0.003 & V=0.966;A=0.005 & 0.2613 & 0 \\

11/1/2025 0:00 & 2 & 0.9893 & -0.02109 & 246.206 & 125.294 & 0 & 0.962 & 53.085 & 0 & 50.0207 & 290.85 & 44.638 & 0.02418 & 0 & 0 & -0.003887 & 0.51455 & 0.43194 & 1 & 0.01358 & 0.11263 & 0 & 0.94718 & V=0.989;A=-0.021 & V=0.994;A=-0.022 & 0.42562 & 0 \\

11/1/2025 0:00 & 3 & 0.971 & 0.06553 & 193.269 & 65.634 & 0 & 17.834 & 59.282 & 0 & 50.0306 & 202.156 & 49.166 & 0.03388 & 0 & 0 & -0.010446 & 0.51438 & 0.57104 & 3 & -0.132 & 0.1704 & 3 & 0.92732 & V=0.971;A=0.066 & V=0.981;A=0.041 & 0.57664 & 0 \\

11/1/2025 0:00 & 4 & 0.9801 & 0.00312 & 167.844 & 86.839 & 0 & 9.995 & 64.967 & 0 & 50.0262 & 156.416 & 42.877 & 0.02633 & 0 & 0 & 0.003364 & 0.48504 & 0.44433 & 3 & 0.04642 & 0.29388 & 1 & 0.94305 & V=0.980;A=0.003 & V=0.977;A=0.003 & 0.44501 & 0 \\

11/1/2025 0:00 & 5 & 0.9346 & 0.01741 & 75.274 & 25.357 & 0 & 13.814 & 59.93 & 0 & 50.0299 & 60.391 & 46.467 & 0.03311 & 0 & 0 & 0.008718 & 0.47516 & 0.46584 & 0 & -0.01688 & 0.16152 & 0 & 0.92912 & V=0.935;A=0.017 & V=0.939;A=0.010 & 0.49499 & 0 \\

11/1/2025 0:00 & 6 & 1.0168 & -0.04082 & 76.907 & 38.062 & 0 & 24.015 & 64.705 & 0 & 50.0343 & 57.272 & 53.307 & 0.02865 & 0 & 0 & -0.04259 & 0.50178 & 0.42589 & 1 & -0.15237 & 0.26443 & 1 & 0.93737 & V=1.017;A=-0.041 & V=1.014;A=-0.028 & 0.48838 & 0 \\

11/1/2025 0:00 & 7 & 0.9813 & 0.06477 & 64.392 & 24.608 & 0 & 9.317 & 68.327 & 0 & 50.0324 & 58.237 & 55.825 & 0.0183 & 0 & 0 & -0.035043 & 0.50515 & 0.49809 & 1 & -0.06264 & 0.22167 & 2 & 0.95781 & V=0.981;A=0.065 & V=0.983;A=0.075 & 0.42269 & 0 \\

11/1/2025 0:00 & 8 & 1.0001 & 0.01637 & 228.018 & 47.125 & 0 & 9.663 & 66.233 & 0 & 50.0283 & 204.872 & 54.154 & 0.02281 & 0 & 0 & -0.030186 & 0.5143 & 0.45037 & 0 & -0.00735 & 0.20201 & 1 & 0.94896 & V=1.000;A=0.016 & V=1.001;A=0.022 & 0.47208 & 0 \\

11/1/2025 0:00 & 9 & 0.9855 & 0.0012 & 173.99 & 36.945 & 0 & 15.049 & 78.187 & 0 & 50.0241 & 167.292 & 43.184 & 0.03086 & 0 & 0 & -0.005805 & 0.49426 & 0.37861 & 1 & 0.01767 & 0.31132 & 3 & 0.93396 & V=0.985;A=0.001 & V=0.985;A=-0.021 & 0.36824 & 0 \\

11/1/2025 0:00 & 10 & 1.048 & 0.07885 & 187.987 & 91.07 & 0 & 4.206 & 69.916 & 0 & 50.0253 & 167.642 & 55.125 & 0.02443 & 0 & 0 & -0.024099 & 0.48806 & 0.67537 & 1 & -0.10852 & 0.1283 & 0 & 0.94563 & V=1.048;A=0.079 & V=1.045;A=0.076 & 0.69615 & 0 \\

11/1/2025 0:00 & 11 & 1.0092 & 0.07002 & 45.518 & 11.869 & 0 & 2.845 & 70.043 & 0 & 50.031 & 52.429 & 55.033 & 0.02997 & 0 & 0 & 0.027609 & 0.47706 & 0.36805 & 2 & -0.23513 & 0.21623 & 3 & 0.93455 & V=1.009;A=0.070 & V=1.004;A=0.052 & 0.35325 & 0 \\

11/1/2025 0:00 & 12 & 0.9493 & -0.06822 & 249.966 & 124.456 & 0 & 5.559 & 71.152 & 0 & 50.018 & 240.692 & 53.811 & 0.02845 & 0 & 0 & 0.030079 & 0.49555 & 0.50997 & 0 & 0.00507 & 0.17544 & 3 & 0.93772 & V=0.949;A=-0.068 & V=0.945;A=-0.061 & 0.52753 & 0 \\

11/1/2025 0:00 & 13 & 1.0314 & -0.0404 & 225.164 & 69.05 & 0 & 4.802 & 64.633 & 0 & 50.0289 & 241.495 & 52.057 & 0.03131 & 0 & 0 & -0.015717 & 0.48217 & 0.42897 & 2 & -0.06389 & 0.12426 & 0 & 0.93217 & V=1.031;A=-0.040 & V=1.026;A=-0.033 & 0.44497 & 0 \\

11/1/2025 0:00 & 14 & 0.9691 & -0.06156 & 88.663 & 21.539 & 0 & 24.62 & 60.613 & 0 & 50.0349 & 63.53 & 44.957 & 0.02964 & 0 & 0 & 0.009086 & 0.51883 & 0.34971 & 3 & -0.11401 & 0.1451 & 2 & 0.93622 & V=0.969;A=-0.062 & V=0.967;A=-0.067 & 0.42003 & 0 \\

11/1/2025 0:00 & 15 & 1.0195 & -0.01104 & 79.128 & 46.212 & 0 & 19.417 & 60.593 & 0 & 50.0286 & 62.604 & 47.123 & 0.03489 & 0 & 0 & 0.024479 & 0.52201 & 0.36721 & 3 & -0.0082 & 0.24288 & 1 & 0.92552 & V=1.019;A=-0.011 & V=1.016;A=-0.025 & 0.40793 & 0 \\

11/1/2025 0:00 & 16 & 0.9966 & -0.03561 & 89.783 & 33.663 & 0 & 18.887 & 50.612 & 0 & 50.0343 & 77.518 & 50.009 & 0.02634 & 0 & 0 & 0.037785 & 0.50503 & 0.3853 & 0 & -0.03165 & 0.23599 & 2 & 0.94232 & V=0.997;A=-0.036 & V=0.996;A=-0.021 & 0.37146 & 0 \\

11/1/2025 0:00 & 17 & 0.9889 & 0.02493 & 103.529 & 35.279 & 0 & 35.702 & 57.671 & 0 & 50.0303 & 63.48 & 47.377 & 0.02866 & 0 & 0 & -0.003871 & 0.50395 & 0.49507 & 0 & 0.01191 & 0.20469 & 1 & 0.93794 & V=0.989;A=0.025 & V=0.985;A=0.023 & 0.43868 & 0 \\

11/1/2025 0:00 & 18 & 0.9923 & 0.02829 & 152.404 & 75.535 & 0 & 21.053 & 49.029 & 0 & 50.0249 & 153.91 & 51.738 & 0.01857 & 0 & 0 & -0.024588 & 0.50201 & 0.4693 & 0 & 0.08931 & 0.18082 & 3 & 0.95769 & V=0.992;A=0.028 & V=0.986;A=0.025 & 0.46619 & 0 \\

11/1/2025 0:00 & 19 & 1.0305 & 0.05653 & 138.099 & 53.425 & 0 & 4.731 & 42.819 & 0 & 50.0281 & 151.209 & 41.186 & 0.02559 & 0 & 0 & 0.000963 & 0.50847 & 0.46155 & 2 & 0.05785 & 0.3563 & 0 & 0.9447 & V=1.031;A=0.057 & V=1.032;A=0.060 & 0.51326 & 0 \\

11/1/2025 0:00 & 20 & 1.0502 & 0.01731 & 94.091 & 31.121 & 0 & 15.33 & 46.746 & 0 & 50.0243 & 96.019 & 43.538 & 0.03056 & 0 & 0 & 0.00071 & 0.51069 & 0.53177 & 0 & -0.082 & 0.15499 & 0 & 0.93453 & V=1.050;A=0.017 & V=1.044;A=0.018 & 0.55388 & 0 \\

11/1/2025 0:00 & 21 & 1.0112 & -0.10024 & 170.76 & 57.929 & 0 & 9.183 & 38.859 & 0 & 50.0316 & 186.919 & 47.462 & 0.03243 & 0 & 0 & 0.010391 & 0.46121 & 0.55942 & 1 & 0.05228 & 0.22122 & 2 & 0.9304 & V=1.011;A=-0.100 & V=0.998;A=-0.099 & 0.60355 & 0 \\

11/1/2025 0:00 & 22 & 1.0062 & -0.05086 & 72.651 & 26.342 & 0 & 18.208 & 32.761 & 0 & 50.028 & 60.646 & 51.347 & 0.02542 & 0 & 0 & 0.010607 & 0.49187 & 0.57467 & 2 & -0.11343 & 0.18942 & 3 & 0.94402 & V=1.006;A=-0.051 & V=1.000;A=-0.038 & 0.45824 & 0 \\

11/1/2025 0:00 & 23 & 1.0733 & 0.15478 & 104.015 & 36.689 & 0 & 25.331 & 32.398 & 0 & 50.0245 & 93.969 & 46.363 & 0.02471 & 0 & 0 & -0.003189 & 0.5055 & 0.37148 & 3 & 0.12174 & 0.16926 & 3 & 0.94594 & V=1.073;A=0.155 & V=1.068;A=0.141 & 0.39793 & 0 \\
11/1/2025 0:00 & 24 & 1.0137 & -0.02318 & 112.853 & 49.938 & 0 & 11.532 & 29.543 & 0 & 50.0298 & 94.112 & 46.472 & 0.02241 & 0 & 0 & -0.015765 & 0.4869 & 0.56775 & 0 & 0.16547 & 0.19975 & 0 & 0.95054 & V=1.014;A=-0.023 & V=1.012;A=-0.022 & 0.66009 & 0 \\

11/1/2025 0:00 & 25 & 0.9699 & -0.02092 & 143.944 & 64.059 & 0 & 3.075 & 32.76 & 0 & 50.0253 & 134.283 & 46.038 & 0.03645 & 0 & 0 & 0.014649 & 0.50678 & 0.45183 & 0 & 0.11513 & 0.22657 & 3 & 0.92249 & V=0.970;A=-0.021 & V=0.965;A=-0.039 & 0.457 & 0 \\

11/1/2025 0:00 & 26 & 0.9306 & -0.00061 & 208.71 & 101.429 & 0 & 7.648 & 22.291 & 0 & 50.032 & 185.069 & 41.981 & 0.03295 & 0 & 0 & -0.022017 & 0.50047 & 0.41468 & 3 & -0.03696 & 0.15654 & 2 & 0.9299 & V=0.931;A=-0.001 & V=0.936;A=-0.009 & 0.38843 & 0 \\

11/1/2025 0:00 & 27 & 0.9279 & -0.05503 & 86.367 & 40.338 & 0 & 17.474 & 27.893 & 0 & 50.0337 & 67.932 & 45.11 & 0.02034 & 0 & 0 & -0.00522 & 0.51277 & 0.38393 & 0 & -0.02177 & 0.23986 & 2 & 0.9548 & V=0.928;A=-0.055 & V=0.931;A=-0.055 & 0.31953 & 0 \\

11/1/2025 0:00 & 28 & 0.9887 & 0.0098 & 154.927 & 92.888 & 0 & 14.445 & 36.179 & 0 & 50.0238 & 144.664 & 41.435 & 0.02338 & 0 & 0 & 0.018177 & 0.50472 & 0.51462 & 2 & -0.12203 & 0.22585 & 3 & 0.9491 & V=0.989;A=0.010 & V=0.980;A=0.015 & 0.51217 & 0 \\

11/1/2025 0:00 & 29 & 0.9973 & -0.00204 & 164.418 & 88.447 & 0 & 34.972 & 33.872 & 0 & 50.0247 & 128.887 & 53.168 & 0.03199 & 0 & 0 & 0.007399 & 0.49998 & 0.55528 & 0 & -0.01938 & 0.27539 & 2 & 0.93071 & V=0.997;A=-0.002 & V=0.998;A=-0.015 & 0.55121 & 0 \\

11/1/2025 0:00 & 30 & 1.0014 & -0.03565 & 55.862 & 28.419 & 0 & 0.127 & 38.414 & 0 & 50.0249 & 63.79 & 48.119 & 0.03284 & 0 & 0 & 0.032398 & 0.52934 & 0.54971 & 2 & -0.0854 & 0.10825 & 1 & 0.9295 & V=1.001;A=-0.036 & V=1.002;A=-0.048 & 0.3982 & 0 \\

11/1/2025 0:00 & 31 & 0.9523 & 0.06983 & 170.029 & 98.837 & 0 & 22.779 & 40.033 & 0 & 50.0312 & 149.821 & 49.571 & 0.02723 & 0 & 0 & 0.009611 & 0.48917 & 0.57121 & 0 & 0.11853 & 0.20181 & 3 & 0.94059 & V=0.952;A=0.070 & V=0.948;A=0.066 & 0.59759 & 0 \\

11/1/2025 0:00 & 32 & 0.9734 & -0.01412 & 89.776 & 31.105 & 0 & 16.39 & 44.338 & 0 & 50.0277 & 70.637 & 46.008 & 0.02621 & 0 & 0 & -0.012894 & 0.5323 & 0.52921 & 3 & -0.01669 & 0.18562 & 0 & 0.94299 & V=0.973;A=-0.014 & V=0.974;A=-0.008 & 0.55216 & 0 \\

11/1/2025 0:00 & 33 & 1.0214 & 0.03409 & 61.61 & 18.293 & 0 & 9.3 & 42.84 & 0 & 50.0263 & 52.538 & 48.732 & 0.02915 & 0 & 0 & -0.024285 & 0.46592 & 0.47287 & 3 & 0.12012 & 0.09131 & 3 & 0.93683 & V=1.021;A=0.034 & V=1.019;A=0.028 & 0.39278 & 0 \\

\bottomrule
\end{tabular}
\end{adjustbox}
}
\label{table1}
\end{table*}

\section{Simulation and Results}
\label{sec:Simulation}

This section presents the simulation study conducted on the IEEE 33-bus radial distribution system to evaluate the performance of the proposed hybrid AI-enabled cyber–physical adaptive control framework. The IEEE 33-bus feeder is widely adopted as a benchmark for optimization, control, and resilience assessment due to its radial topology, nonlinear loading characteristics, and sensitivity to disturbances. The simulation emulates realistic distribution grid behavior under stochastic load fluctuations, renewable intermittency, cyber–physical disturbances, and adversarial events. A total of 6000 discrete time steps were simulated, incorporating measurement noise, communication delays, packet drops, and false data injection (FDI) attacks. To provide an initial understanding of the cyber--physical operating conditions of the test distribution network, Table~\ref{tab:system_data} presents a detailed snapshot of all major system measurements and control variables at the timestamp e.g. 11/1/2025 00:00. This tabulated dataset includes bus voltage and angle information, real and reactive load demand, renewable generation from solar and wind units, battery state-of-charge, EV charging load, feeder power flow, and system frequency. Relevant cyber-layer parameters such as communication delay, packet loss rate, and FDI-attack indicators are also reported. Additionally, Table~\ref{table1} summarizes the outputs of the learning-enabled controllers (ADP, PPO, and DQN), the corresponding resilience index, and both edge-estimated and cloud-optimized system states. This comprehensive operating snapshot serves as the baseline reference for the dynamic simulation results and controller performance analyses presented as follows.

\begin{figure}[!h]
    \centering
    \includegraphics[width=\linewidth]{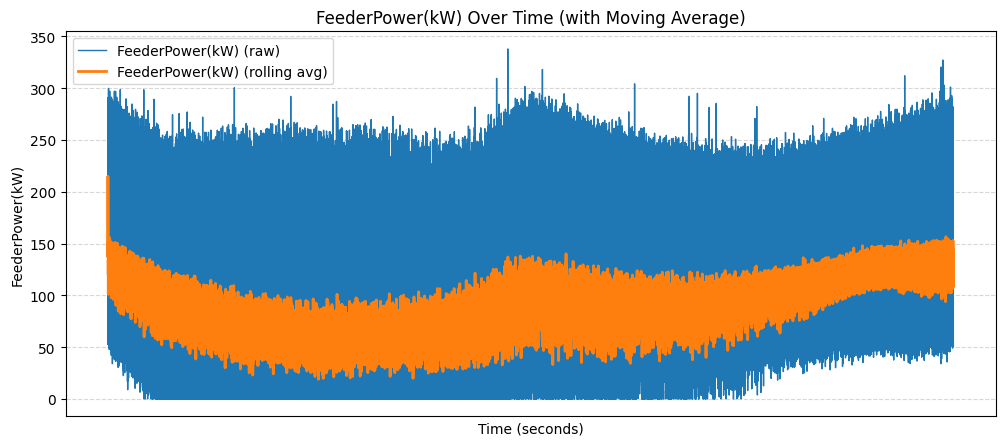}
    \caption{Feeder active power over time (raw and moving average).}
    \label{fig:feeder}
\end{figure}

\begin{figure}[!h]
    \centering
    \includegraphics[width=\linewidth]{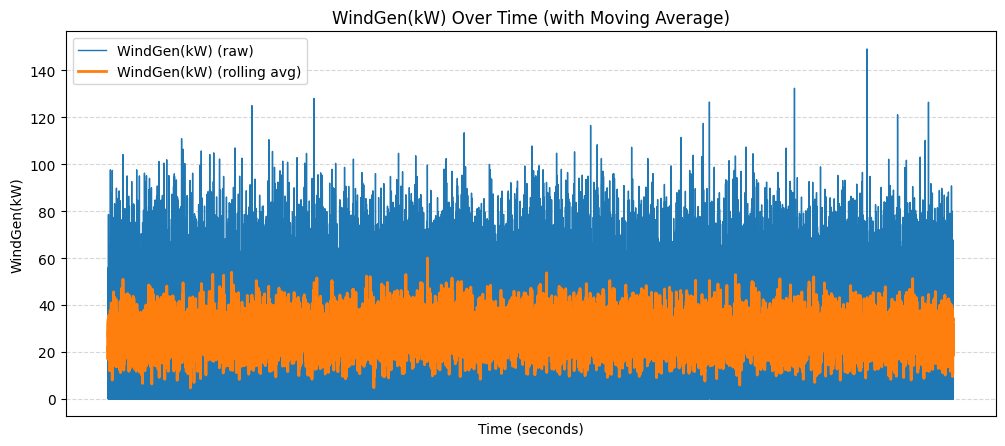}
    \caption{Wind generation profile over time (raw and moving average).}
    \label{fig:wind}
\end{figure}

\begin{figure}[!h]
    \centering
    \includegraphics[width=\linewidth]{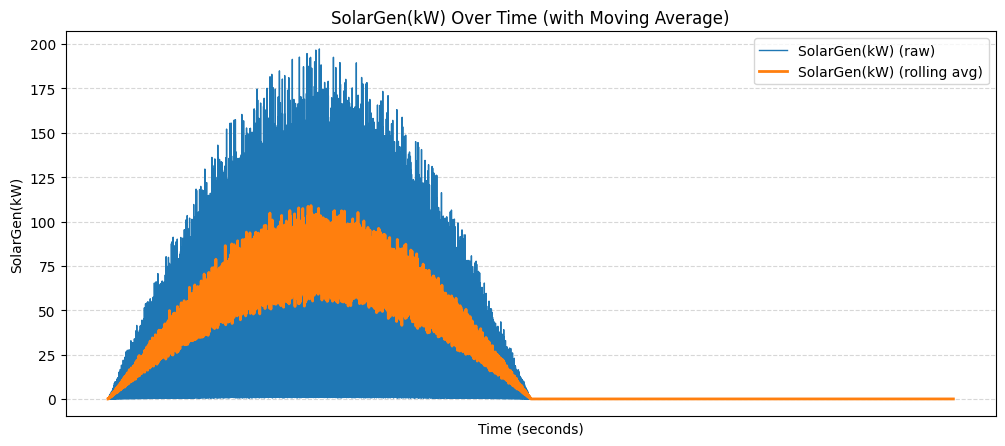}
    \caption{Solar generation profile over time (raw and moving average).}
    \label{fig:solar}
\end{figure}

\begin{figure}[!h]
    \centering
    \includegraphics[width=\linewidth]{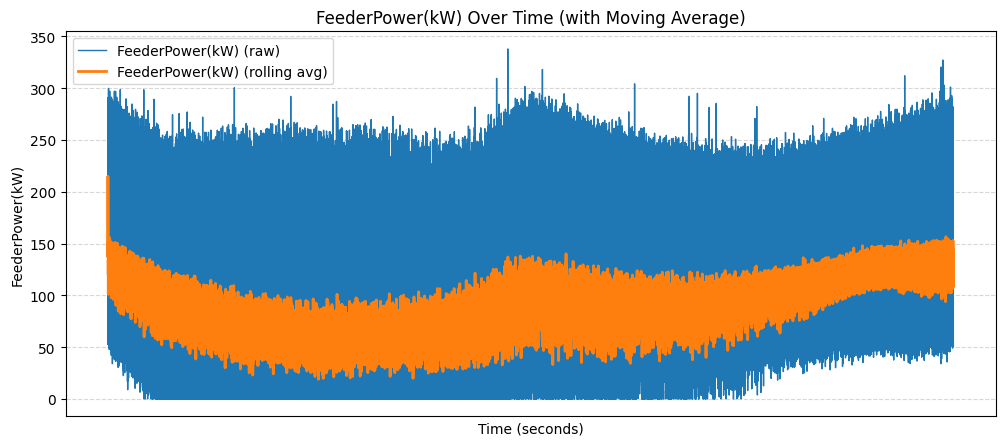}
    \caption{Voltage evolution at Bus 5 with local control actions.}
    \label{fig:voltage}
\end{figure}

\begin{figure}[!h]
    \centering
    \includegraphics[width=\linewidth]{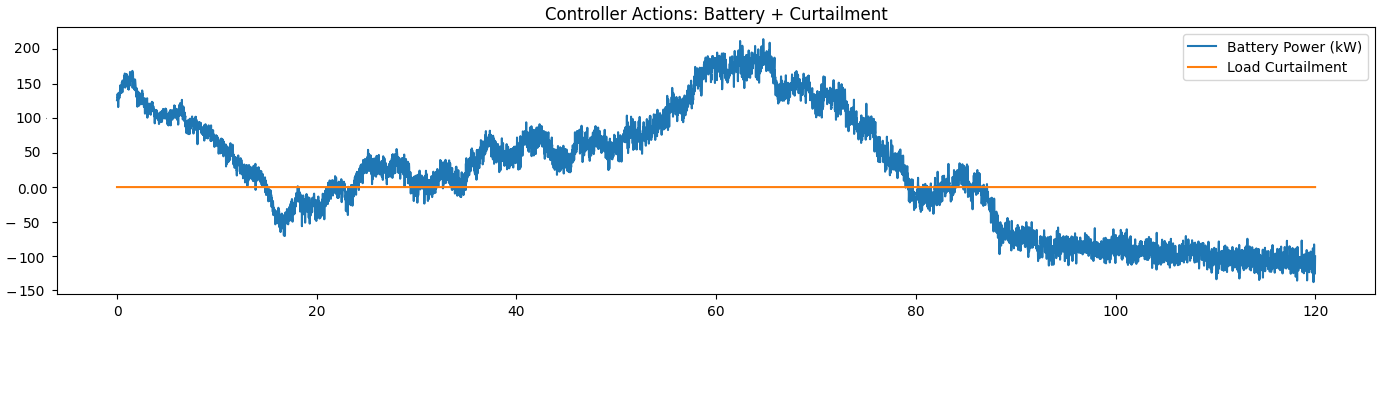}
    \caption{Per-unit Load, PV, and Wind power profile at Bus 5.}
    \label{fig:loadpvwind}
\end{figure}

\begin{figure}[!h]
    \centering
    \includegraphics[width=\linewidth]{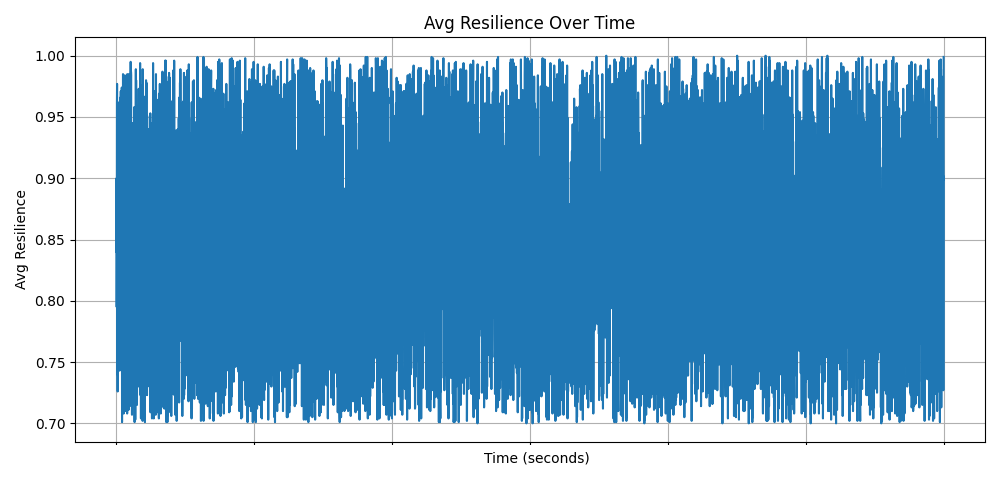}
    \caption{Average resilience index over the simulation horizon.}
    \label{fig:resilience1}
\end{figure}

\begin{figure}[!h]
    \centering
    \includegraphics[width=\linewidth]{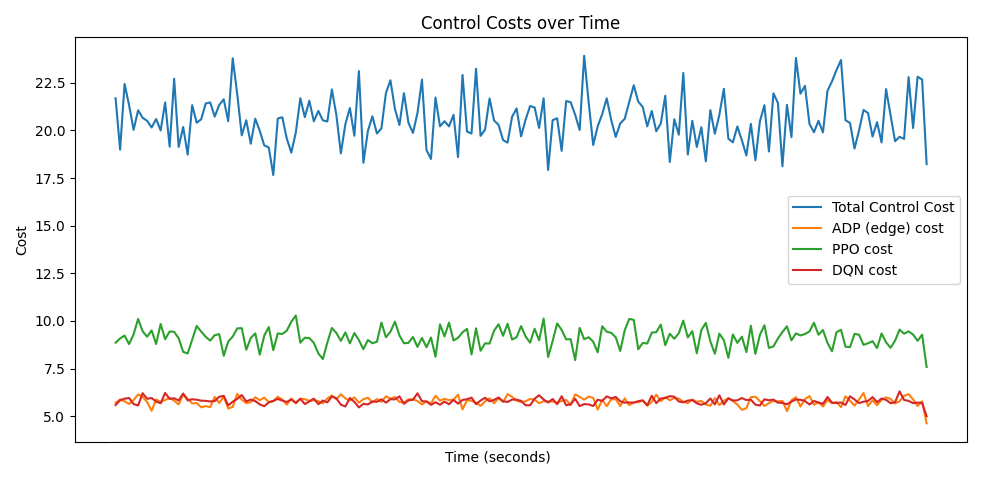}
    \caption{Control cost evolution for Total, ADP, PPO, and DQN controllers.}
    \label{fig:control_costs}
\end{figure}

The hybrid control framework integrates cloud-level PPO with edge-level ADP and DQN agents. The performance is evaluated using multiple indicators including voltage stability, renewable variability response, resilience evolution, and control costs. Figs.~\ref{fig:feeder}--\ref{fig:solar} show raw and smoothed power trajectories for the feeder, load, wind generation, and solar generation. These plots highlight the high-frequency variations in wind, the smooth diurnal pattern of PV, and the joint impact on feeder loading. These disturbances directly influence the learning agents' control actions. 
Fig.~\ref{fig:voltage} displays the voltage evolution at Bus 5 under local control. Disturbances from load, PV, and wind variability cause voltage dips and rises, but the controller maintains voltages close to the nominal 1.0 p.u. band. The minimum voltage remains above 0.94 p.u., demonstrating effective mitigation of congestion and renewable-driven perturbations. 
The battery dispatch and curtailment actions of controller are  shown in Fig.~\ref{fig:loadpvwind}. The battery provides rapid positive and negative power injections to compensate for renewable fluctuations, while load curtailment remains near zero for most of the simulation. Curtailment is only activated briefly during severe disturbance events, confirming that the hybrid control framework minimizes disruptive corrective actions.
\begin{figure}[!h]
    \centering
    \includegraphics[width=\linewidth]{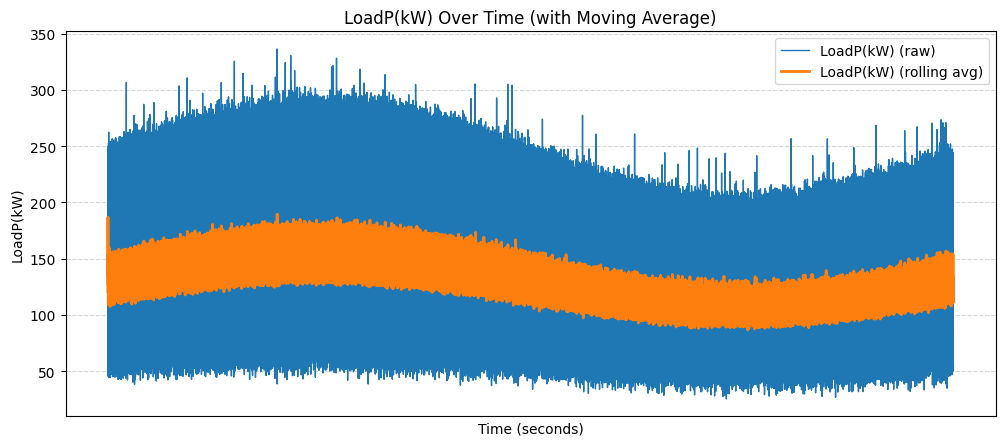}
    \caption{Temporal variation of load showing daily fluctuations and peak/off-peak demand patterns in the smart grid system.}
    \label{fig:controller}
\end{figure}
Fig.~\ref{fig:resilience1} depicts the evolution of the average resilience index. The index remains predominantly in the high-performance band (0.95–1.0), despite occasional dips to 0.70–0.75 caused by rapid load transitions, cloud-induced PV drops, or FDI attacks. Fast recovery after each disturbance demonstrates the system’s strong self-healing and adaptive capability under the hybrid control architecture. 
The updated control cost trajectories are shown in Fig.~\ref{fig:control_costs}. The Total Control Cost fluctuates between 18 and 24, driven primarily by the system-wide corrective actions required during renewable and load disturbances. The PPO cost remains centered around 9, exhibiting stable and smooth behavior indicative of reliable cloud-level policy updates. The ADP and DQN edge controllers show lower costs, typically between 5 and 6, reflecting their fast response and low computational burden. Their trajectories track one another closely, confirming consistent learning behavior across edge devices. Occasional coordinated spikes align with periods of heavy disturbance, but rapid recovery follows each event.
Overall, the hybrid reinforcement learning architecture maintains high operational resilience despite cyber–physical disturbances. The cloud-level PPO agent ensures stable policy learning with minimal variability, while the ADP and DQN edge agents provide robust and fast real-time reactions. The strong coordination between cloud and edge layers results in stable voltages, controlled feeder flows, and low corrective action requirements. The system consistently demonstrates self-healing capability, confirming the suitability of the proposed framework for real-world smart grid applications.

\section{Conclusion}
\label{sec:Conclusion}
The suggested hybrid cyber-physical model of smart grids has been already put into practice, and it proves the success of the 3-layer approach to the smart grid comprising physical, cyber, and control layers. The system has been found to be adaptive, scaled, and even sustainable by incorporating both Adaptive Dynamic Programming (ADP) and AI-based optimization approaches through cloud-independent as well as cloud-assisted contingencies. The use of simulation on a standard IEEE33 bus system has confirmed the fact that the framework is capable of providing grid stability, power dispatch optimization and responding effectively to dynamic operating conditions. By and large, the methodology proves to be an effective and viable solution to the smart grid control and energy management of our current era with the successful implementation. Future development will be to scale the framework to incorporate real time hardware-in-the-loop testing and integration with emerging distributed energy resources to be more resilient and performant.

% \appendices
% \section{Derivation of Swing Equation}
% Appendix details here.

% \section{Proof of Resilience Metric}
% Appendix details here.

% \section*{Acknowledgment}
% This research was supported by [Your Funding Agency]. The authors thank…

\ifCLASSOPTIONcaptionsoff
  \newpage
\fi

\bibliographystyle{IEEEtran}
\bibliography{references}

% Your biographies...
%\begin{IEEEbiography}{Your Name} Biography text. \end{IEEEbiography}

\end{document}